\documentclass[Afour,sageh,times]{sagej}

\usepackage{moreverb,url}

\usepackage[colorlinks,bookmarksopen,bookmarksnumbered,citecolor=red,urlcolor=red]{hyperref}
\usepackage{graphicx}
\usepackage{subfigure}
\usepackage{booktabs} 
\usepackage{amsthm}
\usepackage{amssymb}
\usepackage{bm}
\usepackage{newtxmath}
\usepackage{xcolor}
\usepackage{xspace}
\usepackage{bbm}
\usepackage{algorithm}
\usepackage{algorithmic}
\usepackage[nolist]{acronym}
\usepackage{hyperref}
\usepackage{newtxmath}
\usepackage{dblfloatfix}
\usepackage[shortlabels]{enumitem}

\hypersetup{
    colorlinks=false,
    linkcolor=black,
    filecolor=black,      
    urlcolor=blue,
    }

\colorlet{revision}{black!100}

\newcommand\BibTeX{{\rmfamily B\kern-.05em \textsc{i\kern-.025em b}\kern-.08em
T\kern-.1667em\lower.7ex\hbox{E}\kern-.125emX}}

\usepackage{amsmath,amsfonts,bm}

\def\eqref#1{equation~\ref{#1}}
\def\Eqref#1{Equation~(\ref{#1})}

\def\1{\bm{1}}

\def\vtheta{{\bm{\theta}}}

\def\vc{{\bm{c}}}
\def\vd{{\bm{d}}}

\def\vf{{\bm{f}}}
\def\vg{{\bm{g}}}

\def\vl{{\bm{l}}}

\def\vp{{\bm{p}}}
\def\vq{{\bm{q}}}

\def\vu{{\bm{u}}}

\def\vx{{\bm{x}}}

\def\vz{{\bm{z}}}

\def\mA{{\bm{A}}}

\def\mD{{\bm{D}}}

\def\mG{{\bm{G}}}
\def\mH{{\bm{H}}}
\def\mI{{\bm{I}}}
\def\mJ{{\bm{J}}}
\def\mK{{\bm{K}}}
\def\mL{{\bm{L}}}

\def\mW{{\bm{W}}}

\DeclareMathAlphabet{\mathsfit}{\encodingdefault}{\sfdefault}{m}{sl}
\SetMathAlphabet{\mathsfit}{bold}{\encodingdefault}{\sfdefault}{bx}{n}

\DeclareMathOperator*{\argmin}{arg\,min}

\DeclareMathOperator{\sign}{sign}

\def\vtheta{{\bm{\theta}}}

\def\vpsi{{\bm{\psi}}}
\def\vphi{{\bm{\phi}}}
\def\vtau{{\bm{\tau}}}
\newcommand\norm[1]{\left\lVert#1\right\rVert}

\begin{document}

\runninghead{Lutter and Peters}

\title{Combining Physics and Deep Learning to learn Continuous-Time Dynamics Models}

\author{Michael Lutter\affilnum{1} and Jan Peters\affilnum{1}}

\affiliation{\affilnum{1}Intelligent Autonomous Systems Group at Technical University of Darmstadt, Germany}

\corrauth{Michael Lutter, michael@robot-learning.de}
\begin{acronym}
\acro{hjb}[HJB]{Hamilton-Jacobi-Bellman}
\acro{hji}[HJI]{Hamilton-Jacobi-Isaacs}
\acro{sim2real}[Sim2Real]{simulation to real}
\acro{rl}[RL]{reinforcement learning}
\acro{cfvi}[cFVI]{Continuous Fitted Value Iteration}
\acro{rfvi}[rFVI]{Robust Fitted Value Iteration}
\acro{mlp}[MLP]{Multi-Layer Perceptron}
\acro{delan}[DeLaN]{Deep Lagrangian Networks}
\acro{hnn}[HNN]{Hamiltonian Neural Networks}
\acro{vpt}[VPT]{valid prediction time}
\acro{ode}[ODE]{ordinary differential equation}
\end{acronym}

\begin{abstract}
Deep learning has been widely used within learning algorithms for robotics. One disadvantage of deep networks is that these networks are black-box representations. Therefore, the learned approximations ignore the existing knowledge of physics or robotics. Especially for learning dynamics models, these black-box models are not desirable as the underlying principles are well understood and the standard deep networks can learn dynamics that violate these principles.
%
To learn dynamics models with deep networks that guarantee physically plausible dynamics, we introduce physics-inspired deep networks that combine first principles from physics with deep learning. We incorporate Lagrangian mechanics within the model learning such that all approximated models adhere to the laws of physics and conserve energy. \acf{delan} parametrize the system energy using two networks. The parameters are obtained by minimizing the squared residual of the Euler-Lagrange differential equation. 
%
Therefore, the resulting model does not require specific knowledge of the individual system, is interpretable, and can be used as a forward, inverse, and energy model. Previously these properties were only obtained when using system identification techniques that require knowledge of the kinematic structure. 
We apply \acs{delan} to learning dynamics models and apply these models to control simulated and physical rigid body systems. The results show that the proposed approach obtains dynamics models that can be applied to physical systems for real-time control. Compared to standard deep networks, the physics-inspired models learn better models and capture the underlying structure of the dynamics. 
\end{abstract}

\maketitle

\section{Introduction}
During the last five years, deep learning has shown the potential to fundamentally change the use of learning in robotics. Currently, many robot learning approaches involve a deep network as part of their algorithm. The network either represents a policy that selects the actions, a dynamics model that predicts the next state, or a state estimator that extracts the relevant features from unstructured observations. Initially, many of these approaches were only applicable to simulated environments due to the large amounts of data required to train the networks. When using massive parallelized simulations, these methods achieved astonishing results~\citep{heess2017emergence}. By now, these learning algorithms have been improved and start to be applied to real-world systems~\citep{OpenAI19RubicsCube, haarnoja2018soft}. On the physical systems, the deep network approaches have not bypassed classical robotics techniques yet, but have shown very promising results achieving comparable results as classical methods.

\medskip \noindent
Within many proposed algorithms deep networks have replaced analytic models and other function approximators due to their simplicity, generic applicability, scalability, high model capacity and widespread availability of GPU's enabling fast training and evaluation. The generic applicability of these black-box models combined with the high model capacity is a curse and blessing. On the one side, this combination enables the learning of arbitrary functions with high fidelity. However, this combination is also susceptible to overfit to the data without retrieving the underlying structure. Furthermore, the black-box nature of standard deep networks prevents including prior knowledge from first-order principles. This limitation is especially problematic for robotics as the overfitting to spurious data can lead to unpredictable behaviors damaging the physical system. The problem is also made unnecessarily harder as all existing knowledge of robotics and mechanics is ignored.

\medskip \noindent
In this article, we propose a new approach that combines existing knowledge with deep networks. This combination enables to learn better representations for robotics and retains the advantages of deep networks. To learn physically plausible continuous-time dynamics models of rigid body systems, we combine Lagrangian mechanics with deep networks. The proposed \acf{delan} use two deep networks to parameterize the kinetic and potential energy~\citep{lutter2018deep}. The network parameters are learned by minimizing the squared residual of the Euler-Lagrange differential equation. The resulting dynamics models are guaranteed to evolve as a mechanical system and conserve the system energy. Therefore, these models achieve better long-term predictions and control performance than the standard black-box models. The resulting physics-inspired models share many of the characteristics of analytic models without requiring specific knowledge about the individual system. For example, \ac{delan} models are interpretable and enable the computation of the gravitational forces, the momentum, and the system energy. Previously, computing this decomposition was only possible using the analytic models with the system parameters. \ac{delan} also enables the computation of the forward and inverse models with the same parameters. These characteristics are in stark contrast to standard black-box models. Such black-box models only obtain either the forward or the inverse model and cannot compute the different physical quantities as these need to be learned unsupervised. Due to these advantages of physics-inspired dynamics models, many variants have been proposed \citep{greydanus2019hamiltonian, gupta2019general, zhong2019symplectic, saemundsson2020variational, cranmer2020lagrangian}. 

\subsection{Contribution} \textcolor{revision}{
The contribution of this article is the presentation of a model learning framework that combines the existing knowledge of mechanics with deep networks. To highlight the possibilities of this approach for learning dynamics model, we describe \acf{delan}~\citep{lutter2018deep}. This model learning approach combines deep learning with Lagrangian mechanics to learn a physically plausible model by minimizing the residual of the Euler-Lagrange ordinary differential equation. In contrast to our previous papers \citep{lutter2018deep, Lutter2019Energy}, which mainly focused on specific algorithmic ideas, this article  
\begin{enumerate} [wide=0pt]
    \item consolidates the existing literature on physics-inspired model learning which has been introduced since the initial presentation of \ac{delan}. We summarize the individual contributions and merge the variants into a single big picture.
    \item extends the previous experimental evaluation and provides in-depths comparisons of the different variants of physics-inspired networks. We evaluate the control performance of the learned models on the physical system using inverse dynamics control and energy control. In addition, the performance is compared to system identification and black-box model learning.
    \item provides an elaborate discussion on the current shortcomings of physics-inspired networks and highlight possibilities to overcome these limitations.
\end{enumerate} }

\subsection{Outline}
To provide a self-contained overview about physics-inspired deep networks for learning dynamics models, we briefly summarize the related work~(Section \ref{sec:related_work}), prior approaches for learning dynamics models of rigid body systems as well as the basics of Lagrangian and Hamiltonian mechanics (Section \ref{sec:Preliminaries}). Subsequently, we introduce physics-inspired networks derived from Lagrangian and Hamiltonian mechanics as well as the existing variants (Section \ref{sec:physics_models}). Section \ref{sec:experiments} presents the experimental results of applying these models to model-based control and compares the performance to system identification as well as deep network dynamics models. Finally, Section \ref{sec:discussion} discusses the experimental results, highlights the limitations of physics-inspired networks, and summarizes the contributions of this article.

\section{Related Work}\label{sec:related_work}
In the main part of this article, we focus on learning continuous-time dynamics models of mechanical systems. However, physics-inspired networks and continuous-time deep networks have been utilized for different applications areas. In this section, we want to briefly summarize the existing work on both topics outside the domain of rigid body systems and their differences. 

\subsection{Physics-Inspired Deep Networks}
Incorporating knowledge of physics within deep networks has been approached by introducing conservation laws or symmetries within the network architecture. Both approaches are tightly coupled due to Noether's theorem showing that symmetries induce conservation laws. In the case of conservation laws, these laws can be incorporated by minimizing the residual of the corresponding differential equation to obtain the optimal network parameters. The combination of deep learning and differential equations has been well known for a long time and investigated in more abstract forms \citep{lee1990neural, meade1994solution, lagaris1998artificial, lagaris2000neural}. Using this approach, various authors proposed to use the Navier-Stokes equation~\citep{raissi2017physics_1, chu2021learning}, Schroedinger equation~\citep{raissi2017physics_2}, Burgers Equation~\citep{holl2020learning}, Hamilton's equation~\citep{greydanus2019hamiltonian, zhong2019symplectic, chen2019symplectic, toth2019hamiltonian} or the Euler-Lagrange equation~\citep{lutter2018deep, qin2020machine, cranmer2020lagrangian, gupta2019general}.

\medskip\noindent
Symmetries can be integrated within the network architecture by selecting a non-linear transformation that is either equivariant, i.e., preserves the symmetry, or is invariant to specific transformations of the input. Using this approach, one can derive layers that are translational-, rotational-, scale- and gauge equivariant~\citep{cohen2016group, bekkers2019b, wang2020incorporating, cohen2019gauge}. These architectures are frequently used for computer vision as image classification is translational and rotational invariant~\citep{cohen2019gauge, weiler2019general, lenc2015understanding}. Up to now, only very few papers have applied this approach to model physical phenomena \citep{wang2020incorporating, anderson2019cormorant}. A different approach to symmetries was proposed by \cite{huh2020time}. To obtain time translation invariance, which is equivalent to conservation of energy, this work optimized time-reversibility. Therefore, the symmetry is not incorporated in the network architecture but the optimization loss.

\medskip\noindent
Besides these generic approaches utilizing symmetries and conservation laws, various authors also proposed specific architectures for individual problems. In this case, the known spatial structure of the problem is embedded within the network architecture. For example, \cite{wang2020towards} proposed a network architecture for turbulent flow predictions that incorporates multiple spatial and temporal scales. \cite{sanchez2018graph} used a graph network to encode the known kinematic structure and the local interactions between two links within the network structure. Similarly, \cite{schutt2017quantum} incorporates the local structure of molecules within the network architecture. 

\subsection{Continuous-Time Models \& Neural ODEs}
The work on neural \ac{ode} by \cite{chen2018neural} initiated a large surge of research on continuous-time models. The original work on neural \ac{ode} proposed a deep network with infinite depth to improve classification and density estimation. While these algorithms were not meant for modeling dynamical systems, the explicit integration step within the neural \ac{ode} led to rediscover continuous-time models for dynamical systems. Since then, neural ode's have been frequently mentioned as inspiration to learn continuous-time models~\citep{saemundsson2020variational, huh2020time, botev2021priors, hochlehnert2021learning}. Frequently the term neural \ac{ode} is used interchangeably for a continuous-time model with a deep network. In this work, we will only use the term continuous-time model. One technical difference between the original neural \ac{ode} and continuous-time models is that the neural \ac{ode} uses a variable time step integrator, most commonly the Dormand–Prince method. The continuous-time models use a fixed time step integrator. For the fixed time step integrator, different authors have used the explicit Euler, the Runge Kutta method, or symplectic integrators. For dynamics models, the fixed time step is convenient as the data is observed at a fixed time step determined by the sampling rate of digital sensors. 

\begin{figure*}[t]
    \centering
    \includegraphics[width=\textwidth]{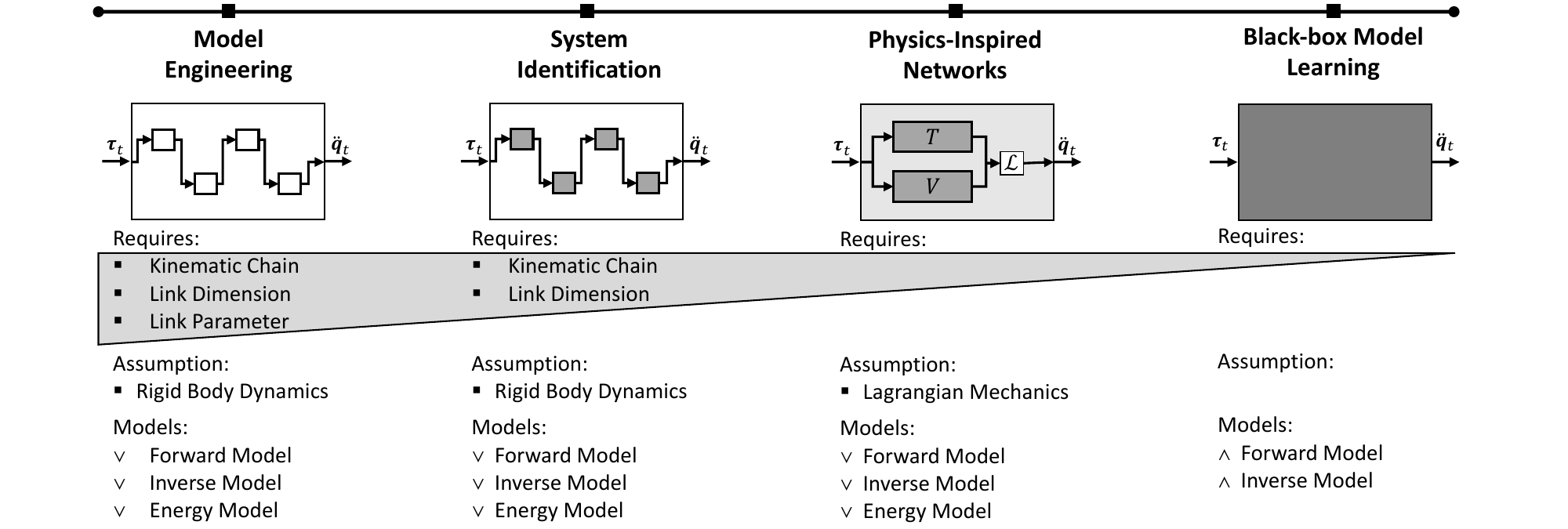}
    \vspace{-1.em}
    \caption{The requirements and assumptions of the different approaches to obtain the dynamics model of mechanical systems. The physics-inspired networks bridge the gap between classical system identification and black-box model learning. While system identification requires knowledge of the kinematic chain, the physics-inspired networks do not require any knowledge of the specific system but obtain comparable characteristics as system identification. Physics-inspired networks also guarantee energy-conserving models and obtain the forward, inverse, and energy model simultaneously.}
    \label{fig:approaches}
\end{figure*} 

\section{Preliminaries}
We want to introduce the standard model learning techniques for dynamical systems and briefly summarize the relevant theory of Lagrangian and Hamiltonian Mechanics. 

\subsection{Learning Dynamics Models} 
Models describing system dynamics, i.e. the coupling of the system input $\vu$ and system state $\vx$, are essential for model-based control approaches \citep{ioannou1996robust} and model based planning. Depending on the approach, one either relies on the forward or inverse model. For example, inverse dynamics control \citep{de2012theory} uses the inverse model to compensate system dynamics, while model-predictive control \citep{camacho2013model} and optimal control \citep{zhou1996robust} use the forward model to compute the future states given an action sequence. For discrete time models, the forward model $f$ maps from the system state $\vx_t$ and input $\vu_t$ to the next state $\vx_{t+1}$. The inverse model $f^{-1}$ maps from system state and the next state to the system input. Mathematically this is described by
\begin{align}
 f(\vx_t, \vu_t; \:\vtheta) &= \vx_{t+1}, &
 f^{-1}(\vx_t, \vx_{t+1}; \: \vtheta) &= \vu_t,
\end{align}
with the model parameters $\vtheta$. In the continuous time setting the next state $\vx_{t+1}$ is replaced with the change of the state $\dot{\vx}$, i.e.,
\begin{align}
 f(\vx_t, \vu_t; \:\vtheta) &= \dot{\vx}_t, &
 f^{-1}(\vx_t, \dot{\vx}_t; \: \vtheta) &= \vu_t.
\end{align}
The continuous-time system can be combined with an integrator, e.g., explicit Euler, Runge-Kutta method, or symplectic integrators, to predict the next state instead of the change of the system state. Therefore, the continuous-time model is independent of the time discretization. Depending on the chosen representation, the transfer function~$f$ and parameters~$\vtheta$ are obtained using different approaches. In the following, we will differentiate the different approaches, (1)~model engineering, (2)~data-driven system identification, and (3)~black-box model learning. In Section~\ref{sec:physics_models}, we will extend these existing categories to physics-inspired models.

\subsubsection{\textbf{Model Engineering.}}
The most classical approach is model engineering, which is predominantly used within the industry. In this case, the transfer function $f$ is the equations of motion and the model parameters are the physical parameters of the robot consisting of the masses, center of gravity, length, and inertia. The equations of motion have to be manually derived for each system. Frequently, one assumes perfect rigid bodies connected by ideal joints and uses Newtonian, Lagrangian or Hamiltonian mechanics and the known structure of the systems to derive the equations. The model parameters can be either inferred using the CAD software or measured by disassembling the individual system. The latter is more precise as it incorporates the deviations due to the manufacturing process \citep{albu2002regelung}. Furthermore, the parameters are identical for the forward and inverse model. Therefore, this approach yields the forward and inverse model simultaneously. To summarize, this approach can yield very precise forward and inverse models for rigid body systems but is labor-intensive as the parameters need to be manually inferred. 

\subsubsection{\textbf{Data-Driven System Identification.}}
Similar to model engineering, data-driven system identification uses the analytic equations of motions as the transfer function. However, the model parameters are learned from observed data rather than measured. Therefore, the equations of motions need to be manually derived but the model parameters are learned. In 1985 four different groups showed concurrently that the dynamics parameters can be obtained by linear regression using hand-designed features for rigid body kinematic chains~\citep{khosla1985parameter, mukerjee1985self, atkeson1986estimation, gautier1986identification}. This approach is commonly referenced as the standard system identification technique for robot manipulators by the textbooks~\citep{siciliano2016springer}. However, this approach cannot guarantee physically plausible parameters as the dynamics parameters \textcolor{revision}{have additional constraints}. For example, this approach can yield negative masses, an inertia matrix that is not positive definite, or violate the parallel axis theorem \citep{ting2006bayesian}. The disadvantages of this approach are that one can only infer linear combinations of the dynamics parameters, cannot apply it to close-loop kinematics \citep{siciliano2016springer} and can only be applied inverse dynamics. The inverse dynamic formulation is problematic as the inverse dynamics do not necessarily have a unique solution due to friction~\citep{ratliff2016doomed}. To overcome these shortcomings, \cite{ting2006bayesian} proposed a projection-based approach while many others~\citep{traversaro2016identification, wensing2017linear, ledezma2017first, sutanto2020encoding, lutter2020differentiable, lutter2021differentiable, geist2021structured} used virtual parametrizations that guarantee physical plausible parameters. For the latter, the optimization does not simplify to linear regression but can be solved using gradient descent. To summarize, this approach only requires the equations of motions analytically and can learn the dynamical parameters from data. Therefore, this approach is not as labor-intensive as model engineering but one needs to ensure to collect 'good' data for the learning.

\subsubsection{\textbf{Black-Box Model Learning.}}
While the previous approaches required knowledge about the individual kinematic chain to derive the equations of motion, the black-box approaches do not require any knowledge of the system. These approaches use any black-box function approximator as a transfer function and optimize the model parameters to fit the observed data. For example, the existing literature used Local Linear Models \citep{schaal2002scalable, haruno2001mosaic}, Gaussian Mixture Models \citep{calinon2010learning, khansari2011learning}, Gaussian Processes \citep{kocijan2004gaussian, nguyen2009model, nguyen2010using, romeres2019derivative, romeres2016online, camoriano2016incremental}, Support Vector Machines \citep{choi2007local, ferreira2007simulation}, feedforward- \citep{jansen1994learning, lenz2015deepmpc,ledezma2017first,sanchez2018graph} or recurrent neural networks \citep{rueckert2017learning, ha2018world, hafner2019dream, hafner2019learning} to learn the dynamics model. The black-box models obtain either the forward or inverse model and the learned model is only valid on the training data distribution. The previous methods based on the analytic equations of motions obtained both models simultaneously and generalize beyond the data distribution as the learned physical parameters are globally valid. However, the black-box models do not require assumptions about the systems and can learn systems including contacts. These previous approaches relied on assuming rigid body dynamics and could only learn the system dynamics of articulated bodies using reduced coordinates without contacts. Therefore, black-box models can be more accurate for real-world systems where the underlying assumption is not valid but is limited to the training domain and rarely extrapolate.

\subsection{Lagrangian Mechanics} \label{sec:Preliminaries}
One approach to derive equations of motion is Lagrangian Mechanics. In the following, we summarize this approach as we will use it in Section~\ref{sec:physics_models} to propose a physics-inspired network for learning dynamics models. More specifically we use the Euler-Lagrange formulation with non-conservative forces and generalized coordinates. For more information and the formulation using Cartesian coordinates please refer to the textbooks \citep{greenwood2006advanced, de2012theory, Featherstone:2007:RBD:1324846}. Generalized coordinates $\vq$ are coordinates that uniquely define the system configuration without constraints. These coordinates are often called reduced coordinates. For articulated bodies, the system state can be expressed as $\vx = \left[\vq, \dot{\vq}\right]$. The Lagrangian mechanic's formalism defines the Lagrangian $\mathcal{L}$ as a function of generalized coordinates $\vq$ describing the complete dynamics of a given system. The Lagrangian is not unique and every $\mathcal{L}$ which yields the correct equations of motion is valid. 
The Lagrangian is generally chosen to be
\begin{align} \label{eq:lagrangian}
\mathcal{L}(\vq, \dot{\vq}) = T(\vq, \dot{\vq}) - V(\vq)
= \frac{1}{2} \dot{\vq}^{\top} \mathbf{H}(\vq) \dot{\vq} - V(\vq),
\end{align}
with the kinetic energy $T$, the potential energy $V$ and the mass matrix $\mathbf{H}(\vq)$. The kinetic energy $T$ is quadratic for any choice of generalized coordinates and any non-relativistic system. The mass matrix is the symmetric and positive definite \citep{de2012theory}. The positive definiteness ensures that all non-zero velocities lead to positive kinetic energy. Applying the calculus of variations yields the Euler-Lagrange equation with non-conservative forces described by
\begin{align} \label{eq:lagrangian_mechanics}
\frac{d}{dt} \frac{\partial \mathcal{L}(\vq, \dot{\vq})}{\partial \dot{\vq}} - \frac{\partial \mathcal{L}(\vq, \dot{\vq})}{\partial \vq} &= \bm{\tau}, \\
\frac{\partial^{2} \mathcal{L}(\vq, \dot{\vq})}{\partial^{2} \dot{\vq}} \ddot{\vq} + \frac{\partial \mathcal{L}(\vq, \dot{\vq})}{\partial \vq \partial \dot{\vq}} \dot{\vq} - \frac{\partial \mathcal{L}(\vq, \dot{\vq})}{\partial \vq} &= \bm{\tau}, \label{eq:hessian_euler_lagrange}
\end{align}
where $\bm{\tau}$ are generalized forces frequently corresponding to the system input $\vu$. Substituting $\mathcal{L}$ with the kinetic and potential energy into \Eqref{eq:lagrangian_mechanics} yields the second order \ac{ode} described by
\begin{align} \label{eq:lagrangian_equality}
\mathbf{H}(\vq) \ddot{\vq} + \underbrace{\dot{\mathbf{H}}(\vq) \dot{\vq} - \frac{1}{2}  \left( \dot{\vq}^{\top} \frac{\partial \mathbf{H}(\vq)}{\partial \vq} \dot{\vq} \right)^{\top}}_{=\:\mathbf{c}(\vq, \dot{\vq})} + \: \underbrace{\frac{\partial V(\vq)}{\partial \vq}}_{=\:\mathbf{g}(\vq)} = \bm{\tau},
\end{align}
where $\mathbf{c}$ describes the forces generated by the Centripetal and Coriolis forces and $\vg$ the gravitational forces \citep{Featherstone:2007:RBD:1324846}. Most robotics textbooks abbreviate this equation as $\mH(\vq)\ddot{\vq} + \vc(\vq, \dot{\vq}) + \vg(\vq) = \vtau$. Using this \ac{ode}, any multi-particle mechanical system with holonomic constraints can be described. Various authors used this \ac{ode} to manually derive the equations of motion for coupled pendulums \citep{greenwood2006advanced}, robotic manipulators with flexible joints \citep{book1984recursive, spong1987modeling}, parallel robots \citep{miller1992lagrange, geng1992dynamic, liu1993singularities} or legged robots \citep{1102105, 1101650}. 

\subsection{Hamiltonian Mechanics}
A different approach to deriving the equations of motions is Hamiltonian mechanics. In this case, the system dynamics are described using the state $\vx = \left[\vq, \vp \right]$ with generalized momentum $\vp$ instead of the generalized velocity $\dot{\vq}$ and the Hamiltonian $\mathcal{H}$ instead of the Lagrangian. The generalized momentum can be expressed using the Lagrangian and is described by $\vp = \partial \mathcal{L} / \partial \dot{\vq}$ \citep{fitzpatrick2008newtonian}. Given the parametrization of the Lagrangian (\Eqref{eq:lagrangian}), this definition is equivalent to $\vp = \mH(\vq) \: \dot{\vq}$. The Hamiltonian describes the complete energy of the system and is defined as 
\begin{align}  \label{eq:hamiltonian}
\mathcal{H}(\vq, \vp) &= T(\vq, \vp) + V(\vq)
= \frac{1}{2} \vp^{\top} \mH(\vq)^{-1} \vp + V(\vq). 
\end{align}
The Hamiltonian can be computed by applying the Legendre transformation to the Lagrangian which is described by
\begin{align}
  \mathcal{H}(\vq, \dot{\vq}) = \dot{\vq}^{\top} \frac{\partial \mathcal{L}(\vq, \dot{\vq})}{\partial \dot{\vq}} - \mathcal{L}(\vq, \dot{\vq}).
\end{align}
Using the generalized momentum $\vp$ and the generalized coordinate $\vq$, the Euler-Lagrange equation can be rewritten to yield Hamilton's equations with control \citep{greenwood2006advanced}. Hamilton's equations is described by
\begin{gather}
    \dot{\vq} = \frac{\partial \mathcal{H}(\vq, \vp)}{\partial \vp}, \hspace{30pt} \dot{\vp} = - \frac{\partial \mathcal{H}(\vq, \vp)}{\partial \vq} + \vtau. \label{eq:hamiltons_equation}
\end{gather}
The Euler-Lagrange equation (\Eqref{eq:lagrangian_equality}) can be easily derived from Hamilton's equation by substituting \Eqref{eq:hamiltonian} into \Eqref{eq:hamiltons_equation} and using the definition of the generalized momentum, 
i.e.,
\begin{gather*}
\dot{\vp} = - \frac{\partial \mathcal{H}(\vq, \vp)}{\partial \vq} + \vtau, \\
\frac{d}{dt} \Big[ \mH(\vq) \, \dot{\vq} \Big] = \frac{1}{2} \left(\vp^{\top} \mH^{-1} \frac{\partial \mH}{\partial \vq} \mH^{-1} \vp \right)^{\top} - \frac{\partial V}{\partial \vq} + \vtau, \\
\mH \ddot{\vq} + \dot{\mH} \dot{\vq} = \frac{1}{2} \left(\dot{\vq}^{\top} \frac{\partial \mH}{\partial \vq} \dot{\vq} \right)^{\top} - \frac{\partial V}{\partial \vq} + \vtau, \\
\mH \ddot{\vq} + \dot{\mH} \dot{\vq} - \frac{1}{2} \left(\dot{\vq}^{\top} \frac{\partial \mH}{\partial \vq} \dot{\vq} \right)^{\top} + \frac{\partial V}{\partial \vq} = \vtau.
\end{gather*}
Many textbooks omit the generalized forces within Hamilton's equation but adding these generalized forces is straightforward as shown in the previous derivation.

\section{Physics-Inspired Deep Networks}  \label{sec:physics_models}
\textcolor{revision}{A different approach to black-box model learning is to combine black-box models with physics to guarantee a physically plausible dynamics model. 
One combination is to use deep networks to represent the system energy and use the resulting Lagrangian to derive the equations of motion using the Euler-Lagrange differential equation. This approach was initially proposed by \cite{lutter2018deep} with the presentation of Deep Lagrangian Networks (DeLaN). Since then, many papers exploring variations of these approaches have been proposed including approaches that use Hamiltonian mechanics instead of Lagrangian mechanics \citep{greydanus2019hamiltonian, cranmer2020lagrangian, gupta2019general, zhong2019symplectic, sanchezgonzalez2019hamiltonian, saemundsson2020variational, zhong2021differentiable, zhong2020dissipative, hochlehnert2021learning}.} 

\medskip\noindent
\textcolor{revision}{All of these models have in common, that the learned dynamics models conserve energy when the non-conservative forces can be modeled properly and the generalized coordinates are observed. Therefore, the learned model is guaranteed to adhere to one of the fundamental concepts of physics. This property is beneficial as it has been shown within prior research that naive deep network dynamics models frequently increase or decrease the system energy even when the energy should be conserved \citep{greydanus2019hamiltonian, zhong2019symplectic, hochlehnert2021learning, saemundsson2020variational}. 
}

\begin{figure*}[t]
    \centering
    \includegraphics[width=\textwidth]{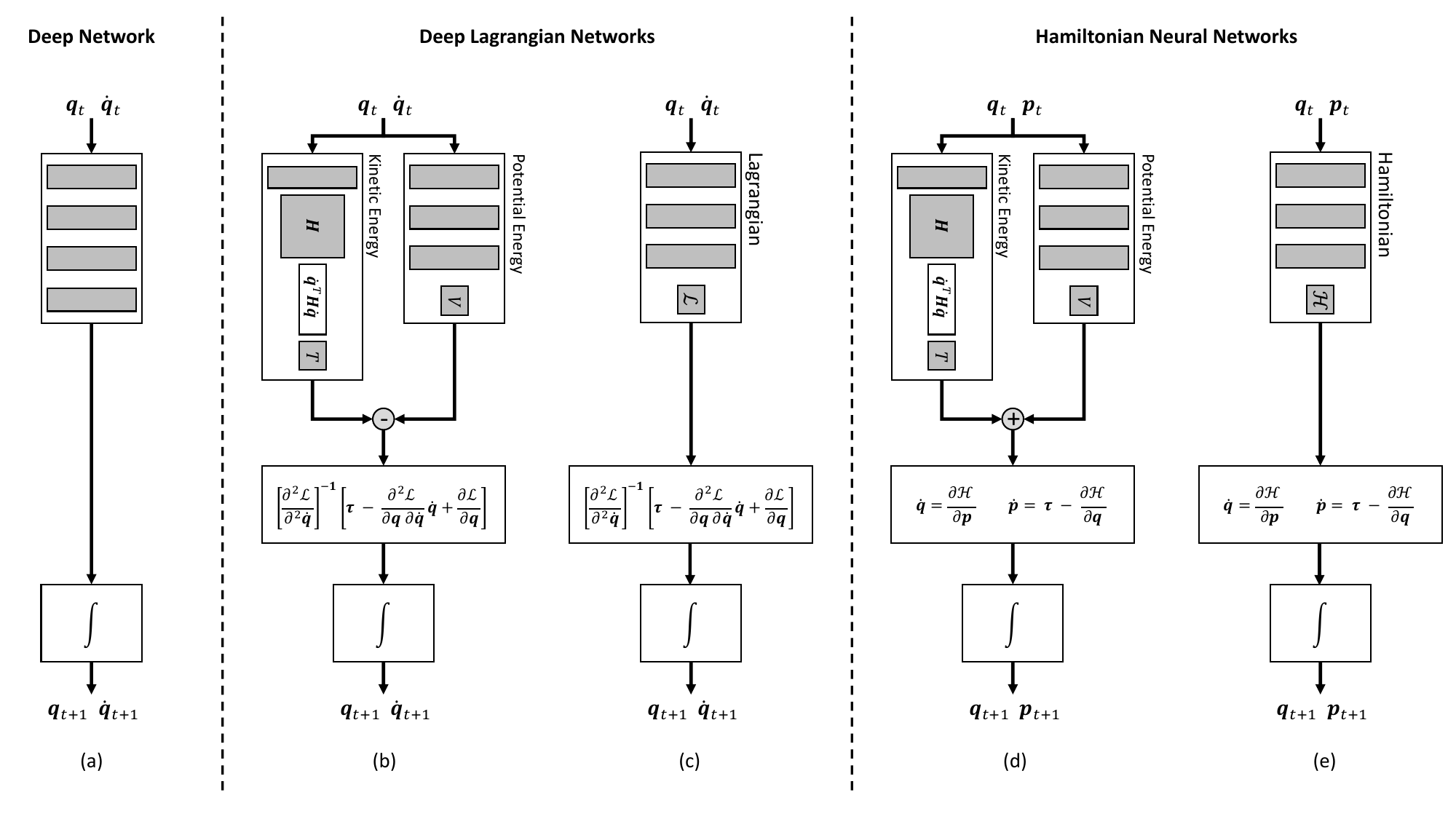}
    \vspace{-1.em}
    \caption{\textcolor{revision}{The flowcharts of a continuous-time forward model using a deep network and the physics-inspired networks forward models. (a) Standard deep model learning approach, where a network is used to directly predict the change in position and velocity. (b-c) Deep Lagrangian Networks which use deep networks to predict the Lagrangian $\mathcal{L}$ of the dynamical system and compute the change in position and velocity using the Euler-Lagrange differential equation. (b) shows the structured Lagrangian approach, where two networks predict the kinetic $T$ and potential energy $V$ and computes the Lagrangian is analytically using $\mathcal{L} = T - V$. (c) shows the black-box Lagrangian approach where a network directly predicts $\mathcal{L}$. (d-e) Hamiltonian Neural networks which use deep networks to predict the Hamiltonian $\mathcal{H}$ and computes the change in position and impulse via Hamilton's equation. Similar to the Lagrangian variants, (d) shows the structured Hamiltonian and (e) the black-box Hamiltonian. The structured Hamiltonian computes the Hamiltonian via $\mathcal{H} = T(\vq, \dot{\vq}) - V(\vq)$, where the kinectic and potential energy is predicted using two networks. The black-box Hamiltonian uses a single network to directly predict $\mathcal{H}$.}}
    \label{fig:flowchart}
\end{figure*} 

\medskip \noindent
In the following, we present \ac{delan} (Section \ref{sec:delan}) and the combination of Hamiltonian mechanics and deep networks in Section \ref{sec:hamilton}. Afterwards, Section \ref{sec:variants} describes all the proposed extensions of \ac{delan} and \ac{hnn}. Therefore, this section provides the big picture of existing physics-inspired deep networks for learning dynamics models. The flow charts of the variants are shown in Figure~\ref{fig:flowchart}

\subsection{Deep Lagrangian Networks (DeLaN)} \label{sec:delan}
\ac{delan} is one instantiation of these physics-inspired deep networks. \ac{delan} parametrizes the mass matrix $\mH$ and the potential energy $V$ as two separate deep networks. Therefore, the approximate Lagrangian $\mathcal{L}$ described by
\begin{align}
    \mathcal{L}(\vq, \dot{\vq};\: \vpsi, \vphi)= \frac{1}{2} \dot{\vq}^{\top} \mH(\vq;\, \vpsi)\, \dot{\vq} - V(\vq; \, \vphi). 
\end{align}
Using this parametrization the forward and inverse model can be derived. The forward model $\ddot{\vq} = f(\vq, \dot{\vq}, \vtau; \: \vpsi, \vphi)$ is described by
\begin{equation*} \label{eq:forward_model}
\begin{aligned}
\ddot{\vq} &=  \bigg[ \frac{\partial^{2} \mathcal{L}(\vq, \dot{\vq})}{\partial^{2} \dot{\vq}} \bigg]^{-1} \bigg[ \vtau - \frac{\partial \mathcal{L}(\vq, \dot{\vq})}{\partial \vq \partial \dot{\vq}} \dot{\vq} + \frac{\partial \mathcal{L}(\vq, \dot{\vq})}{\partial \vq} \bigg], \\
 &= \hspace{22pt} \mH^{-1} \hspace{22pt} \bigg[ \vtau - \dot{\mH} \dot{\vq} + \frac{1}{2}  \left( \dot{\vq}^{\top} \frac{\partial \mathbf{H}}{\partial \vq} \dot{\vq} \right) ^{\top} - \: \frac{\partial V}{\partial \vq} \bigg].
\end{aligned}
\end{equation*}
The inverse model $\vtau = f^{-1}(\vq, \dot{\vq}, \ddot{\vq}; \: \vpsi, \vphi)$ is described by
\begin{equation} \label{eq:inverse_model}
\begin{aligned}
\vtau &= \frac{\partial^{2} \mathcal{L}(\vq, \dot{\vq})}{\partial^{2} \dot{\vq}} \ddot{\vq} + \frac{\partial \mathcal{L}(\vq, \dot{\vq})}{\partial \vq \partial \dot{\vq}} \dot{\vq} - \frac{\partial \mathcal{L}(\vq, \dot{\vq})}{\partial \vq}, \\
&=\mH \ddot{\vq} + \dot{\mH} \dot{\vq} - \frac{1}{2}  \left( \dot{\vq}^{\top} \frac{\partial \mathbf{H}}{\partial \vq} \dot{\vq} \right)^{\top} + \: \frac{\partial V}{\partial \vq}.
\end{aligned}
\end{equation}
The partial derivatives within the forward and inverse model can be computed using automatic differentiation or symbolic differentiation. See \cite{lutter2018deep} for the symbolic differentiation of the mass matrix and the deep networks. 

\medskip\noindent
The system energy cannot be learned using supervised learning as the system energy cannot be observed. Therefore, the network weights of the kinetic and potential energy are learned unsupervised using the temporal consequences of the actions and system energy. One approach to learn the network parameters is to minimize the residual of the Euler-Lagrange differential equation. This optimization problem is described by
\begin{align} \label{eq:opt_residual}
\vpsi^{*}, \vphi^{*} = \argmin_{\vpsi, \vphi} \: \norm{\: \frac{d}{dt} \frac{\partial \mathcal{L}}{\partial \dot{\vq}} - \frac{\partial \mathcal{L}}{\partial \vq_i} - \bm{\tau} \: }_{\mW_{\tau}}^{2},
\end{align}
with the Mahalanobis norm $\norm{\cdot}_{\mW}$ and the diagonal covariance matrix of the generalized forces $\mW_{\tau}$. It is beneficial to normalize the loss using the covariance matrix because magnitude of the residual might vary between different joints. This optimization can be solved using any gradient-based optimization technique. Minimizing the squared residual is equivalent to fitting the inverse model, i.e., $\vpsi^{*}, \vphi^{*} = \argmin_{\vpsi, \vphi} \norm{\vtau - f^{-1}(\vq, \dot{\vq}, \ddot{\vq}; \: \vpsi, \vphi)}_{\mW_{\tau}}^{2}$. This loss can also be extended to include the forward prediction that fits the joint accelerations. The combined optimization problem is described by
\begin{equation} \label{eq:lagrangian_loss}
\begin{aligned}
 \vpsi^{*}, \vphi^{*} = \argmin_{\vpsi, \vphi} \: & \norm{\vtau - f^{-1}(\vq, \dot{\vq}, \ddot{\vq}; \: \vpsi, \vphi)}_{\mW_{\tau}}^{2} \\ 
& \hspace{20pt} +  \: \norm{\ddot{\vq}- f(\vq, \dot{\vq}, \vtau; \: \vpsi, \vphi)}_{\mW_{\ddot{\vq}}}^{2},    
\end{aligned}
\end{equation}
with the diagonal covariance matrix of the generalized forces~$\mW_{\tau}$ and accelerations~$\mW_{\ddot{\vq}}$. Furthermore, its beneficial to add regularization in the form of weight decay as the Lagrangian is not unique. The Euler-Lagrange equation is invariant to linear transformation. Hence, the Lagrangian $\mathcal{L}' = \alpha L + \beta$ solves the Euler-Lagrange equation if $\alpha$ is non-zero and $\mathcal{L}$ is a valid Lagrangian. Therefore, adding weight regularization helps obtaining a unique solution. 

\subsubsection{\textbf{Positive-Definite Mass Matrix.}}
To obtain a physically plausible kinetic energy, the mass matrix has to be positive definite, i.e.,
\begin{align}
    \vq^{\top} \hspace{3pt} \mH(\vq) \: \vq > 0 \hspace{10pt} \forall \hspace{10pt} \vq \in \mathbb{R}_{0}^{n}.
\end{align}
This constraint ensures all non-zero velocities have positive kinetic energy for all joint configurations. We obtain a positive definite mass matrix by predicting the Cholesky decomposition of the mass matrix with a small positive offset $\epsilon$ on the diagonal instead of the mass matrix directly. Therefore, the mass matrix is described by
\begin{align}
    \mH(\vq) = \mL(\vq) \mL(\vq)^{\top} + \epsilon \mI,
\end{align}
with lower triangular matrix $\mL$ and identity matrix $\mI$. In addition, the diagonal needs to be positive as otherwise, the mass matrix is only positive semi-definite. A positive diagonal is ensured by adding a non-negative linearity to the elements of the diagonal and the positive offset $\epsilon$. Using this parametrization the mass matrix is ensured to be positive definite for all joint configurations. However, this parametrization is numerically not favorable \textcolor{revision}{because the} random weights the diagonal is close to $\epsilon$ for all inputs. This small diagonal is problematic for the forward model as the small eigenvalues of the mass matrix lead to a large amplification of the control torques due to the matrix inverse. The default diagonal can be shifted by adding the positive constant $\alpha$ before the non-linearity. This shift is described by
\begin{gather*}
    \vl_{\text{diag}} = \sigma ( \vl_{\text{diag}} + \alpha) + \epsilon,
\end{gather*}
with the vectorized diagonal~$\vl_{\text{diag}}$ and the softplus function~$\sigma$. If $\alpha > 1$, the mass matrix damps the applied torques when $\vl_{\text{diag}} \approx \mathbf{0}$. This transformation is not essential to obtain good results but balances the forward and inverse losses.

\subsubsection{\textbf{Advantages of DeLaN.}}
In contrast to the black-box model, this parametrization of the dynamics has three advantages, (1) this approach yields a physically plausible model that conserves energy, (2) is interpretable, and (3) can be used as forward, inverse, and energy model. The DeLaN model is guaranteed to evolve like a mechanical system and is passive \citep{spong1987modeling} as the forward dynamics are derived from the physics prior and the positive definite mass matrix for all model parameters. If the system is uncontrolled, i.e., $\vtau = \mathbf{0}$, the system energy is conserved as the change in energy is described by $\dot{\mathcal{H}} = \dot{\vq}^{\top} \vtau = \mathbf{0}$. In contrast, black-box models can generate additional energy without system inputs. It is important to note that the conservation of energy of \ac{delan} does not guarantee to prevent the divergence of the model rollouts. The potential energy is not bounded and can accelerate the system indefinitely. Especially outside the training domain, the potential energy is random.

\medskip \noindent
The model is interpretable as one can disambiguate between the different forces, e.g., inertial-, centrifugal-, Coriolis, and gravitational force. This decomposition is beneficial as some model-based control approaches require the explicit computation of the mass matrix, the gravitational force, or the system energy. Furthermore, the same model parameters can be used for the forward, inverse, and energy model. Therefore, the forward and inverse model are consistent. In contrast, black-box models need to learn separate parameters for the inverse and forward model that might not be consistent and cannot obtain the system energy as these cannot be observed. 

\subsection{Hamiltonian Neural Networks (HNN)}\label{sec:hamilton} 
Instead of using Lagrangian Mechanics as model prior for deep networks, \cite{greydanus2019hamiltonian} proposed to use Hamiltonian mechanics. In this case, the \ac{hnn} parametrize the Hamiltonian with two deep networks described by
\begin{align}
    \mathcal{H}(\vq, \vp;\: \vpsi, \vphi)= \frac{1}{2} \vp^{\top} \mH(\vq;\, \vpsi)^{-1}\, \vp - V(\vq; \, \vphi). 
\end{align}
It is important to note that \ac{hnn} predict the inverse of the mass matrix instead of the mass matrix as in \ac{delan}. Similar to \ac{delan}, the forward and inverse model can be derived. The forward model $\left[ \dot{\vq}, \dot{\vp}\right] = f(\vq, \vp, \vtau; \: \vpsi, \vphi)$ is described by
\begin{gather*}
\dot{\vq} =  \hspace{7pt} \frac{\partial \mathcal{H}(\vq, \vp) }{ \partial \vp} =\mH^{-1} \vp, \\
\dot{\vp} = - \frac{\partial \mathcal{H}(\vq, \vp) }{ \partial \vq } + \vtau 
= -\frac{1}{2} \left(\vp^{\top} \frac{\partial \mH^{-1}}{\partial \vq} \vp \right)^{\top} - \frac{\partial V}{\partial \vq} + \vtau.
\end{gather*}
The inverse model $\vtau = f^{-1}(\vq, \vp, \dot{\vp}; \: \vpsi, \vphi)$ is described by
\begin{align*}
\vtau &= \dot{\vp} + \frac{\partial \mathcal{H}(\vq, \vp)}{\partial \vq}, \\
 &= \dot{\vp} - \frac{1}{2} \left(\vp^{\top} \mH^{-1} \frac{\partial \mH}{\partial \vq} \mH^{-1} \vp \right)^{\top} + \frac{\partial V}{\partial \vq}.
\end{align*}
%
%
The network parameters of the kinetic and potential energy can be obtained by minimizing the squared residual using the observed data consisting of $\left[\vq, \vp, \dot{\vq}, \dot{\vp}, \vtau \right]$. This optimization is described by
\begin{equation} \label{eq:hamiltonian_loss}
\begin{aligned}
    \vpsi^{*}, \vphi^{*} = \argmin_{\psi, \phi} \:\:  &\norm{\: \dot{\vp} + \frac{\partial \mathcal{H}(\vq, \vp)}{\partial \vq} - \vtau \:}_{\mW_{\dot{\vp}}}^2 \\
    &\hspace{29pt} + \:\norm{\: \dot{\vq} - \frac{\partial \mathcal{H}(\vq, \vp)}{\partial \vp}\:}_{\mW_{\dot{\vq}}}^{2}, 
\end{aligned}
\end{equation}
with the diagonal covariance matrix $\mW_{\dot{\vq}}$ and $\mW_{\dot{\vp}}$ of $\dot{\vq}$ and $\dot{\vp}$. The minimization can be solved using the standard gradient based optimization toolkit and automatic differentiation.

\subsubsection{\textbf{Differences to DeLaN.}}
\ac{delan} and \ac{hnn} share the same advantages as both models are derived from the same principle. Therefore, \ac{hnn} conserve energy, are interpretable, and provide a forward, inverse, and energy model. The main difference is that \ac{delan} uses position and velocity while \ac{hnn} uses position and momentum. Depending on the observed quantities either model fits better than the other. A minor difference is that minimizing the residual of the Euler-Lagrange equation is identical to the inverse model loss while minimizing the residual of Hamilton's equations is identical to the forward model loss. 

\medskip \noindent
From a numerical perspective, the Hamiltonian mechanics prior is slightly beneficial as the forward and inverse model only relies on the inverse of the mass matrix. Therefore, one does not need to numerically compute the inverse of the predicted matrix. Avoiding the explicit inversion makes the learning and model rollout a bit more stable. The Lagrangian mechanics prior relies on the mass matrix as well on the inverse. Therefore, the inverse of the predicted matrix has to be computed numerically. When the eigenvalues of the mass matrix approach $\epsilon$ and $\epsilon \ll 1$, the model rollout and the optimization of the forward model can become numerically sensitive. Therefore, it is important to choose $\epsilon$ as large as possible for the corresponding system as this limits the amplification of the acceleration.

\subsection{Variations of DeLaN \& HNN} \label{sec:variants} 
Since the introduction of \ac{delan} \citep{lutter2018deep} and HNN~\citep{greydanus2019hamiltonian}, many other variants and extensions have been proposed within the literature. We provide an overview of the existing work and highlight the differences.

\subsubsection{\textbf{Parametrization of $\mathcal{L}$ and $\mathcal{H}$.}} \label{ssec:structure}
In the previous sections, the Hamiltonian~$\mathcal{H}$ and Lagrangian~$\mathcal{L}$ were parameterized by two networks predicting the mass matrix, or its inverse, for the kinetic energy and the potential energy. Instead of predicting these two quantities separately, one can also use a single feed-forward network for both quantities. This factorization is described by
\begin{align*}
    \mathcal{L} &= h(\vq, \dot{\vq}; \: \vpsi), & \mathcal{H} &= h(\vq, \vp; \: \vpsi),
\end{align*}
with the standard feed-forward network $h$ and the network parameters $\vpsi$. Within the literature, this approach was used by \citep{greydanus2019hamiltonian, cranmer2020lagrangian} while the \citep{lutter2018deep, gupta2020structured, zhong2019symplectic, saemundsson2020variational, finzi2020simplifying} used the representation of kinetic and potential energy. \textcolor{revision}{In the following we, will differentiate between both approaches by using the term structured Lagrangian/Hamiltonian and black-box Lagrangian/Hamiltonian. The structured approach, represent the mass matrix and potential energy explicitly while the black-box approach uses a single network to represent the Lagrangian or Hamiltonian. The differences in the model architecture for both approaches are depicted in Figure \ref{fig:flowchart}.}

\medskip\noindent
One benefit of using a black-box $\mathcal{L}$ and $\mathcal{H}$ is that the quadratic parametrization of the kinetic energy does not apply to relativistic systems. The disadvantages of a single network approach are that this parametrization is computationally more demanding as one needs to compute the Hessian of the network. Evaluating the Hessian of a deep network can be done using automatic differentiation, but is expensive in terms of computation and memory. When using the quadratic kinetic energy, computing the Hessian of the network is not needed. Furthermore, the Hessian may not be invertible if only a single network is used. If the Hessian is singular or nearly singular, the forward model using the Lagrangian prior becomes unstable and diverges. For structured Lagrangian, this problem does not occur as the eigenvalues of the mass matrix are lower bounded.  

\medskip\noindent
Most existing work uses standard feed-forward networks to model the system energy, the Hamiltonian or the Lagrangian \citep{lutter2018deep, greydanus2019hamiltonian, zhong2019symplectic, gupta2020structured, saemundsson2020variational, finzi2020simplifying}. Other variants have also applied the physics-inspired networks to graph neural networks \citep{sanchezgonzalez2019hamiltonian, cranmer2020lagrangian, botev2021priors}. Such graph neural networks incorporate additional structure within the network architecture when the system dynamics consist of multiple identical particles without additional constraints. Therefore, these methods exhibit improved performance for modeling N-body problems. 

\subsubsection{\textbf{Loss Functions and Integrators.}} \label{sec:integrator}
The loss functions of \ac{delan} (\Eqref{eq:lagrangian_loss}) and \ac{hnn} (\Eqref{eq:hamiltonian_loss}) express the loss in terms including the acceleration, i.e., $\ddot{\vq}$ and $\dot{\vp}$. These quantities are commonly not observed for real-world systems and are approximated using finite differences. The problem of this approximation is that the finite differences amplify the amplitude of high frequency noise components. Therefore, one has to use low-pass filters to obtain good acceleration estimates. A different approach that avoids approximating the accelerations is to only use the forward loss and reformulate the loss in terms of position and velocities. In this case, the loss is described by 
\begin{align}
\vpsi^{*}, \vphi^{*} = \argmin_{\vpsi, \vphi} \: \norm{\: \vx_{t+1} - \hat{\vx}_{t+1}(\vx_t, \vtau_t; \vpsi, \vphi) \:}^{2},
\end{align}
with the predicted next state $\hat{\vx}$, the state $\vx = \left[ \vq, \dot{\vq} \right]$ in the case of Lagrangian formulation and the state $\vx = \left[ \vq, \vp \right]$ in the Hamiltonian formulation. The predicted next state can be obtained by solving the differential equation
\begin{gather*}
\begin{bmatrix}
\dot{\vq}\\
\ddot{\vq}
\end{bmatrix} =
\begin{bmatrix}
\dot{\vq} \\
\frac{\partial^{2} \mathcal{L}(\vq, \dot{\vq})}{\partial^{2} \dot{\vq}}^{-1} \left[\bm{\tau} - \frac{\partial \mathcal{L}(\vq, \dot{\vq})}{\partial \vq \partial \dot{\vq}} \dot{\vq} + \frac{\partial \mathcal{L}(\vq, \dot{\vq})}{\partial \vq} \right]
\end{bmatrix}, 
\end{gather*}
\begin{gather*}
\begin{bmatrix}
\dot{\vq}\\
\dot{\vp}
\end{bmatrix} =
\begin{bmatrix}
\frac{\partial \mathcal{H}(\vq, \vp) }{ \partial \vp} \\
-\frac{\partial \mathcal{H}(\vq, \vp) }{ \partial \vq }+ \mG(\vq) \: \vtau
\end{bmatrix},
\end{gather*}
using any numerical integration approach. In the case of the explicit Euler integration this approach is identical to the loss of \Eqref{eq:lagrangian_loss} and \Eqref{eq:hamiltonian_loss}. A common choice to compute the next step is the Runge Kutta 4 (RK4) fixed time step integrator~\citep{gupta2019general, greydanus2019hamiltonian}. This loss formulation also enables a multi-step loss which has been shown to improve the performance of model predictive control for deterministic models~\citep{lutter2021learning}

\medskip\noindent
A more elaborate approach has been proposed by \cite{saemundsson2020variational} that combines discrete mechanics with variational integrators. This combination guarantees that even the discrete-time system conserves momentum and energy. The RK4 integration might leak or add energy due to the discrete-time approximation. The main disadvantage of the variational integrator networks is that this approach assumes a constant mass matrix. Therefore, the Coriolis and centrifugal force disappear (\Eqref{eq:lagrangian_equality}) and the acceleration only depends on the position. Within the discrete mechanics literature, extensions exist to apply the variational integrator to multi-body systems with a non-constant mass matrix. However, these extensions are non-trivial and involve solving a root-finding problem within each integration step~\citep{lee2020linear}. 

\subsubsection{\textbf{Feature Transformation.}}
The previous sections always used generalized coordinates or momentum to describe the system dynamics. However, this formulation can be problematic as these coordinates are unknown or unsuitable for function approximation. For example, continuous revolute joints without angular limits are problematic for function approximation due to the wrapping of the angle at $\pm \pi$. This problem is commonly mitigated using sine/cosine feature transformations. Such feature transformation can be included in physics-inspired networks if the feature transforms mapping from the generalized coordinates to the features $\vz$ is known and differentiable. The more general problem of only observing the features and unknown feature transformation and generalized coordinates is discussed in Section \ref{sec:generalized_coordinates}. 

\medskip\noindent
Let $g$ be the feature transform mapping generalized coordinates to the features $\vz = g(\vq)$. For continuous revolute joints the feature transform $\vg(\vq) = \left[\cos{\vq_0}, \sin{\vq_0} \right]$ avoids the problems associated with wrapping the angle. In this case the Lagrangian is described by
\begin{align*}
    \mathcal{L}(\vz, \dot{\vq};\: \vpsi, \vphi)= \frac{1}{2} \dot{\vq}^{\top} \mH(\vz;\, \vpsi)\, \dot{\vq} - V(\vz; \, \vphi). 
\end{align*}
In this case, one can apply the chain rule to obtain the gradients w.r.t. the generalized coordinates, i.e., 
\begin{align*}
    \frac{\partial \mathcal{L}}{\partial \vq} = \frac{\partial \vg}{\partial \vq}^{\top} \frac{\partial \mathcal{L}}{\partial \vz}.
\end{align*}
This approach is identical to adding an input layer to the neural network with the hand-crafted transformations. The feature transformation was previously introduced by \cite{zhong2019symplectic}. However, the authors only manually derived the special case for continuous angle while this approach can be easily generalized to arbitrary differentiable feature transformations. 

\begin{figure*}
  \begin{minipage}[c]{0.73\textwidth}
    \includegraphics[width=\textwidth]{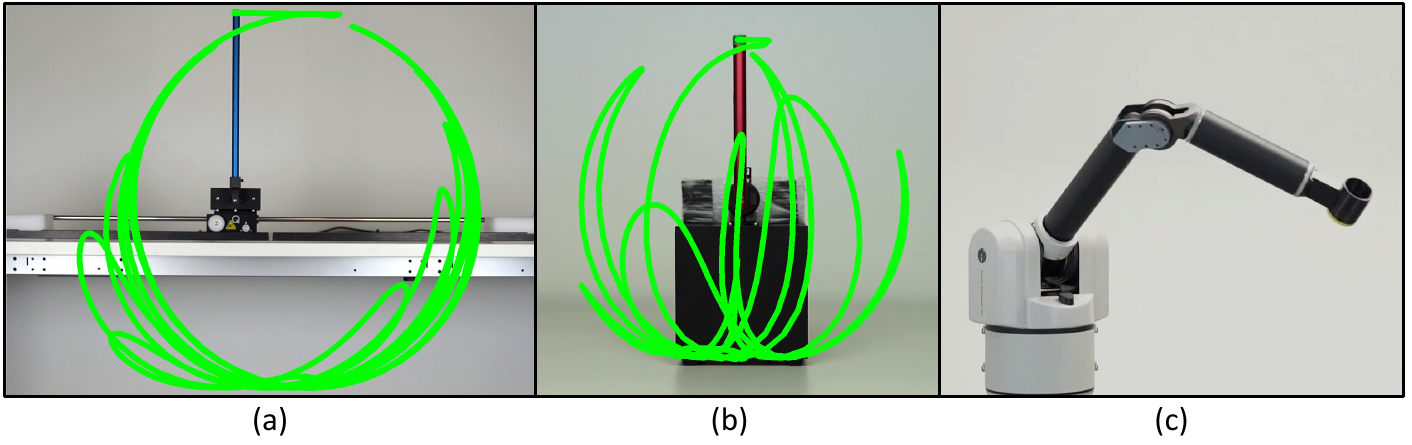}
  \end{minipage}
  \hfill
  \begin{minipage}[l]{0.25\textwidth}
    \caption{The (a) Cartpole, (b) Furuta pendulum and (c) Barrett WAM used for the evaluation. The Furuta pendulum and cartpole perform a swing-up using the energy controller. The Barrett WAM executes a cosine trajectory with a different frequency per joint.  
    } \label{fig:systems}
  \end{minipage}
\end{figure*}

\subsubsection{\textbf{Actuator Models and Friction.}} \label{sec:friction}
The physics-inspired networks cannot model friction directly as the learned dynamics are conservative. Incorporating friction within this model learning approach in a non-black-box fashion is non-trivial because friction is an abstraction to combine various physical effects. For robot arms in free space, the friction of the motors dominates, for mechanical systems dragging along a surface the friction at the surface dominates while for legged locomotion the friction between the feet and floor dominates but also varies with time. Therefore, defining a general case for all types of friction in compliance with the Lagrangian and Hamiltonian Mechanics is challenging. Various approaches to incorporate friction models analytically can be found in \citep{lurie2013analytical, wells1967schaum}.

\medskip\noindent
Most existing works on physics-inspired networks only focus on friction caused by the actuators, which dominates for robot arms~\citep{Lutter2019Energy, gupta2019general, lutter2020differentiable}. In this case the friction can be expressed using generalized coordinates and is a non-conservative force. Incorporating other types of friction than actuator friction is non-trivial as these cannot be easily expressed using the generalized force. In this case, one requires the contact-point and contact Jacobian to map the contact-force to the generalized force. For the actuator model, the generalized force required for \ac{delan} and \ac{hnn} is expressed using an addition function that modulates the system input and adds friction. This function is described by $\vtau = g(\vq, \dot{\vq}, \vu)$ with the system input $\vu$. For the actuator model, one can either choose a white-box approach that uses a analytic actuator and friction models~\citep{Lutter2019Energy} or a black-box approach that uses a deep network~\citep{gupta2019general, zhong2019symplectic}. For example, a white-box model can add a friction torque ${\vtau_f}$ to motor torque $\vu$. Therefore, the generalized force is described $\vtau = \vu + \vtau_{f}$.  Within the literature many friction models have been proposed \citep{olsson1998friction, albu2002regelung, bona2005friction, wahrburg2018motor}. These models assume that the motor friction only depends on the joint velocity $\dot{\vq}_i$ of the $i$th-joint and is independent of the other joints. Common choices for friction are static, viscous, stiction described by
\begin{align*}
    &\text{Coulomb Friction} & 
    \vtau_{f} &= -\vtau_c, \\
    &\text{Viscous Friction} & 
    \vtau_{f} &= - \bm{\rho} \odot \dot{\vq}, \\
    &\text{Stiction} & 
    \vtau_{f} &= -\tau_{s} \odot \sign\left(\dot{\vq} \right) \odot \exp\big(-\dot{\vq}^{2} / \nu  \big), 
\end{align*}
with the elementwise multiplication $\odot$, the Coulomb friction constant $\vtau_{c}$, the viscous friction constant $\bm{\rho}$ and the stiction constants $\vtau_s$ and $\nu$. These friction models can also be combined to yield the Stribeck friction described by
\begin{align*}
    \vtau_{f} &= - \left( \tau_{c} + \tau_{s} \odot  \exp\left(-\dot{\vq}^{2} / \nu  \right) \right) \odot  \text{sign}\left(\dot{\vq} \right) - \vd  \odot \dot{\vq}.
\end{align*}
It is important to note that the system is not time-reversible when stiction is added to the dynamics as multiple motor-torques can generate the same joint acceleration \citep{ratliff2016doomed}.

\medskip\noindent
In contrast to these white-box approaches, \cite{gupta2019general} and \cite{zhong2019symplectic} proposed to add an black-box actuator model. For example, \cite{gupta2019general} proposed to use a black-box control matrix $\mG(\vq)$ with viscous friction for \ac{delan}. Therefore, the actuator model is described by 
\begin{align}
    \vtau = \mG(\vq, \dot{\vq}; \vtheta) \: \vu - \bm{\rho} \odot \dot{\vq},
\end{align}
with the positive friction coefficients $\bm{\rho}$. The control matrix $\mG$ is predicted by an additional neural network. Similarly, \cite{zhong2020dissipative} proposed to use a state-dependent control matrix $\mG(\vq)$ and a positive definite dissipation matrix $\mD(\vq)$ for \ac{hnn}. In this case the generalized force is described by
\begin{gather*}
\vtau = \mG(\vq) \: \vu - \mD(\vq) 
\begin{bmatrix}
\frac{\partial \mathcal{H}(\vq, \vp) }{ \partial \vq} \\
\frac{\partial \mathcal{H}(\vq, \vp) }{ \partial \vp }
\end{bmatrix}.
\end{gather*} 
Both matrices are predicted using a deep network. The network parameters of the actuator model are optimized using gradient descent. These black-box actuator models can represent more complex actuator dynamics and even system dynamics violating the assumptions of Lagrangian and Hamiltonian Mechanics. However, this actuator model can also result that the potential and kinetic energy are ignored and only the black-box model dominates the predicted dynamics. To avoid that the actuator model predicts the complete system dynamics, it is beneficial to add penalties to the magnitude of the actuator during the optimization. The existing grey-box model learning literature \citep{lutter2020differentiable, hwangbo2019learning, allevato2020tunenet} has shown that these penalties improve the performance. 

\begin{table*}[b]
\color{revision}
\footnotesize
\centering
\renewcommand{\arraystretch}{1.4}
\caption{\footnotesize
The hyperparameters of DeLaN used for the different dynamical systems. All physical system have a fixed sampling time of $2.0$ms. The number of training samples for the 2-DoF robot differs by dataset. See table \ref{table:results} for more details about the different datasets.
}
\setlength{\tabcolsep}{22.0pt}
\begin{tabular*}{\textwidth}{| l | c | c | c | c |}
\hline
&  2-DoF Robot & Cartpole & Furuta Pendulum & Barrett WAM\\
%
\hline
Total DoF / Actuated DoF& 
2 / 2&
2 / 1 & 
2 / 1 &
4 / 4
\\ 
Number of Training Samples & 
$2500$ \& $10^{5}$ &
$10^{5}$ / $\approx200$s & 
$10^{5}$ / $\approx200$s &
$10^{5}$ / $\approx200$s
\\ 
Network Dimension & 
[$2$ x $64$] & [$2$ x $256$] &
[$2$ x $128$] & [$2$ x $128$] 
\\ 
Activation & 
Tanh  & SoftPlus & 
SoftPlus & SoftPlus 
\\
Batch Size &  
$512$ & $1024$ & 
$1024$ & $1024$ 
\\
Learning Rate &  
$10^{-4}$ & $10^{-4}$ & 
$10^{-4}$ & $10^{-4}$ 
\\
Weight Decay &  
$10^{-5}$ & $10^{-5}$ & 
$10^{-5}$ & $10^{-5}$ 
\\
Optimizer & 
ADAM & ADAM & 
ADAM & ADAM \\
\hline
\end{tabular*}
\label{table:hyperparameters}
\end{table*}

\section{Experiments} \label{sec:experiments}
In the experiments, we apply physics-inspired deep network models to learn the non-linear dynamics of simulated systems and physical systems. Within the simulation experiments, we want to test whether the different physics-inspired networks learn the underlying structure and highlight the empirical differences of the existing approaches. On the physical systems, we compare the model-based control performance of \ac{delan} with a structured Lagrangian for the fully-actuated and under-actuated system to standard system identification techniques and black-box model learning. We only use \ac{delan} for the physical systems as for these systems we do not observe the momentum. Hence, only the Lagrangian physics prior is applicable. One could treat the Hamiltonian prior as a latent space problem with the momentum being the latent representation. However, this approach would effectively boil down to the Lagrangian prior. Using these experiments, we want to answer the following questions:

\medskip\noindent
\textbf{Q1:} Do physics-inspired networks learn the underlying representation of the dynamical system? 

\medskip\noindent
\textbf{Q2:} Do physics-inspired networks perform better than continuous-time black-box models?

\medskip\noindent
\textbf{Q3:} Can physics-inspired networks be applied to physical systems where the physics prior does not hold?

\begin{figure*}[t]
    \centering
    \includegraphics[width=\textwidth]{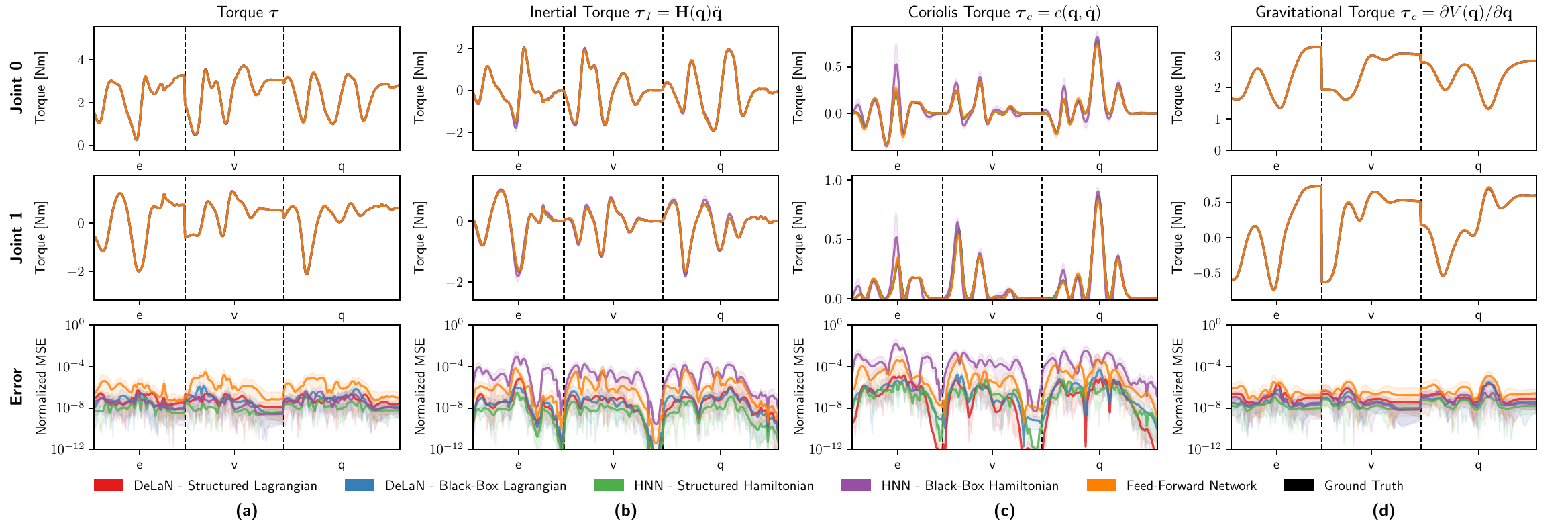}
    \caption{\textcolor{revision}{(a) The learned inverse model using the character dataset averaged over 10 seeds. The test character 'e', 'v', 'q' are not contained within the training set. The remaining columns show predicted force decomposition. (b) plots the inertial force $\mathbf{H}\ddot{\vq}$, (c) the Coriolis and Centrifugal forces $c(\vq, \dot{\vq})$ and (d) the gravitational force $g(\vq)$. All physics-inspired networks learn a good inverse model that obtains a lower MSE than the feed-forward network. The Lagrangian approaches learn a better force decomposition than the Hamiltonian approach. This improved performance is especially visible for the inertial, centrifugal, and Coriolis torque.}}
    \label{fig:inverse_model}
\end{figure*}

\subsection{Experimental Setup}
\textcolor{revision}{To answer these questions, we apply the different variations of physics-inspired models to 4 different systems and compare the performance to three baselines. Within the experiments, we denote the physics-inspired networks that only use a single network to represent the Lagrangian or Hamiltonian as black-box DeLaN/HNN. When two separate networks are used to represent that mass-matrix and potential energy, we refer to this approach as structured DeLaN/HNN. The detailed differences between both approaches are described in section \ref{ssec:structure}. For each of the experiments the dynamics models are learned from a fixed dataset and are trained until convergence. All evaluations are performed on a test dataset that is not contained within the training dataset.}

\medskip\noindent
In the following, we briefly introduce the systems and baselines. The code of \acf{delan} and \acf{hnn} is available at \url{https://github.com/milutter/deep_lagrangian_networks}.

\subsubsection{\textbf{Plants.}} Within the experiments, we apply the model learning techniques to a simulated two-link pendulum, the Barrett WAM, the cartpole, and the Furuta pendulum. For all physical systems, only the joint position is directly observed. The velocities and accelerations are computed using finite differences and low-pass filters. These filters are applied offline to use zero-phase shift filters that do not cause a phase shift within the observations. \textcolor{revision}{The hyperparameters of \ac{delan} for each system are described in Table \ref{table:hyperparameters}.}

\medskip \noindent
\textbf{Two-link Pendulum.} The two-link pendulum has two continuous revolute joints, is fully actuated, and acts in the vertical x-z plane with gravity. The pendulum is simulated using Bullet \citep{coumans2018}.

\medskip \noindent
\textbf{Cartpole.} The physical cartpole (Figure~\ref{fig:systems}a) is an under-actuated system manufactured by \cite{quanser}. The pendulum is passive and the cart is voltage controlled with up to $500$Hz. The linear actuator consists of a plastic cogwheel drive with high stiction. 

\medskip \noindent
\textbf{Furuta Pendulum.} 
The physical Furuta pendulum (Figure~\ref{fig:systems}b) is an under-actuated system manufactured by \cite{quanser}. Instead of the linear actuator of the cartpole, the Furuta pendulum has an actuated revolute joint and a passive pendulum. The revolute joint is voltage controlled with up to $500$Hz. The main challenge with this system is the small masses and length scale of the system. These characteristics yield a very sensitive control system. 

\medskip \noindent
\textbf{Barrett WAM.} The Barrett WAM (Figure~\ref{fig:systems}c) consists of four actuated degrees of freedom controlled via torque control with $500$Hz. The actuators are back-driveable and consist of cable drives with low gear ratios enabling fast accelerations. The joint angles sensing is on the motor-side. Therefore, any deviation between the motor position and joint position due to the slack of the cables cannot be observed. We only use the 4 degree of freedom version as the wrist and end-effector joints cannot be excited due to the limited range of motion and acceleration.

\subsubsection{\textbf{Baselines.}}
We use the analytic dynamics model, system identification, and a feed-forward deep network as baselines. 

\medskip \noindent
\textbf{Analytic Model.} The analytic model uses the equation of motion derived using rigid body dynamics and the system parameters, i.e., masses, center of gravity, and inertias, provided by the manufacturer. In addition to the rigid body dynamics, these models are augmented with a viscous friction model. 

\medskip \noindent
\textbf{System Identification.} This approach requires the knowledge of the analytic equations of motions and infers the system parameters from data. More specifically we use the technique described by \cite{atkeson1986estimation}. This approach showed that for rigid body kinematic trees the inverse dynamics model is a linear model described by 
\begin{gather}
    \vtau = \mA(\vq, \dot{\vq}, \ddot{\vq}) \: \vtheta,
\end{gather}
with the with hand-crafted features $\mA(\cdot)$ derived from the kinematics and the system parameters $\vtheta$. As the inverse dynamics are a linear model, the system parameters can be obtained using linear regression. We additionally penalize deviations from the parameters nominal parameters provided by the manufacturer. In this case the optimal parameters inferred from data are obtained by
\begin{gather}
\vtheta^{*} = \vtheta_{0} + \left( \mA^{\top} \mA + \lambda^{2} \mI \right)^{-1} \mA^{\top} \Big(\vtau - \mA \vtheta_{0} \Big),
\end{gather}
with the nominal parameters $\vtheta_{0}$ and the regularization constant $\lambda$. The resulting system parameters \textcolor{revision}{might} not be physically plausible as the individual elements of $\vtheta$ \textcolor{revision}{have additional constraints}~\citep{ting2006bayesian}. For example the masses have to be positive and the inertias have to adhere to the parallel axis theorem. 

\begin{figure*}[t]
    \centering
    \includegraphics[width=\textwidth]{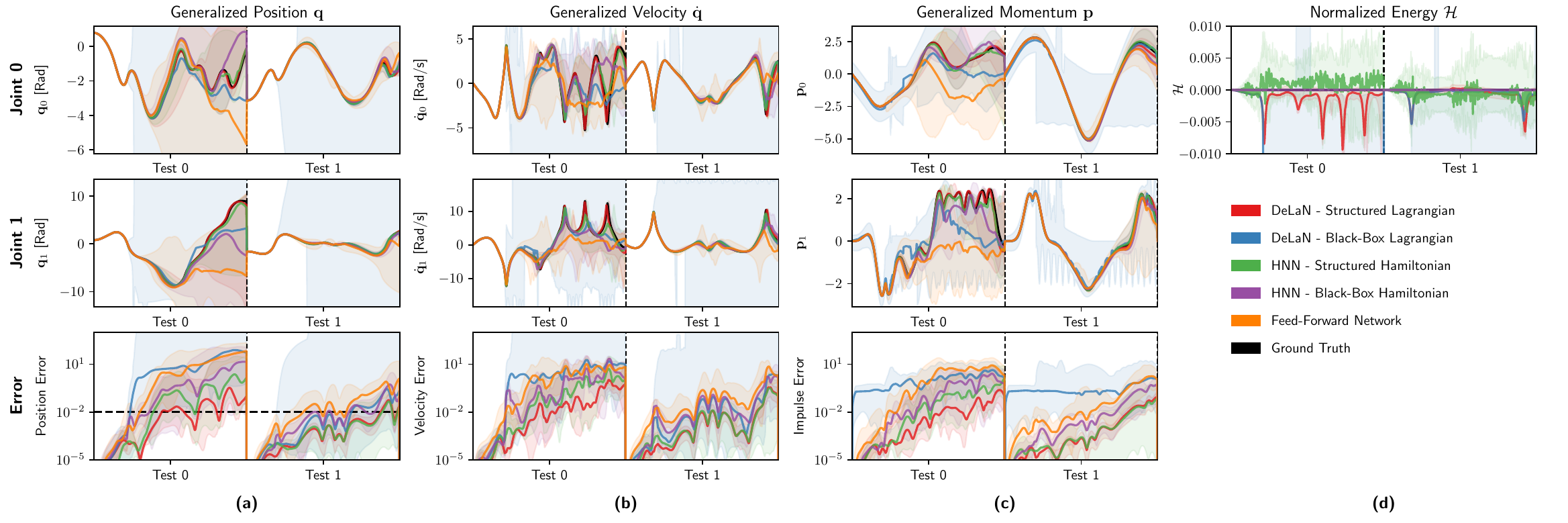}
    \caption{\textcolor{revision}{The model rollouts of (a) the position, (b) velocity and (c) momentum, and (d) energy of the forward models for two test trajectories of the uniform dataset averaged over 10 seeds. The structured physics-inspired networks perform the best compared to the standard feed-forward network and the black-box counterparts. Especially the rollout of the black-box Lagrangian commonly diverges as the Hessian of the Lagrangian, which is required for computing the acceleration, becomes close to singular. This nearly singular Hessian causes exploding velocities and consequently also divergence of the estimated momentum computed via $\mathbf{H} \dot{\vq}$. In the appendix we provide additional plots with only one model per figure to enable an in-depth comparison of the error-bounds.}}
    \label{fig:forward_model}
\end{figure*} 

\begin{table*}[!b]
\scriptsize
\centering
\renewcommand{\arraystretch}{1.25}
\caption{The normalized mean squared error (nmse) and mean \acf{vpt} as well as the corresponding confidence interval averaged over 10 seeds. On average the structured Hamiltonian and Lagrangian approaches obtain better forward and inverse models than the black-box counterparts and the standard feed forward neural network. When observing the corresponding phase space coordinates, the Hamiltonian and Lagrangian approaches perform comparable. 
}
\setlength{\tabcolsep}{5.pt}
\begin{tabular*}{\textwidth}{c c | c c c c | c c }
\toprule
& & \multicolumn{4}{c |}{\textbf{Inverse Model}}  & \multicolumn{2}{c}{\textbf{Forward Model}}  \\
\multicolumn{2}{c|}{\textbf{Uniform Data - $\#$ Samples} $\mathbf{=100,000}$} &
Torque - $\vtau$ & 
Inertial Torque $\vtau_{\mI}$ & 
Coriolis Torque $\vtau_{\vc}$ & 
Gravitational Torque $\vtau_{g}$ & 
State Error $\dot{\vx}$  & 
VPT [s] \\
\cmidrule(lr){1-2} \cmidrule(lr){3-3} \cmidrule(lr){4-4} \cmidrule(lr){5-5} \cmidrule(lr){6-6} \cmidrule(lr){7-7} \cmidrule(lr){8-8} 
DeLaN & Structured Lagrangian & 
$\mathbf{2.2\mathrm{e}{-7} \pm 3.2\mathrm{e}{-6}}$    &     $\mathbf{2.9\mathrm{e}{-9} \pm 4.5\mathrm{e}{-8}}$    &     
$\mathbf{2.5\mathrm{e}{-8} \pm 3.0\mathrm{e}{-7}}$    &     $\mathbf{1.0\mathrm{e}{-8} \pm 3.2\mathrm{e}{-8}}$    &
$\mathbf{3.9\mathrm{e}{-5} \pm 5.9\mathrm{e}{-4}}$    &     $\mathbf{5.64\text{s} \pm 1.78\text{s}}$ \\
DeLaN & Black-Box Lagrangian & 
$2.1\mathrm{e}{-4} \pm 4.9\mathrm{e}{-3}$    &    $\mathbf{4.0\mathrm{e}{-9} \pm 2.1\mathrm{e}{-8}}$    &     
$1.9\mathrm{e}{-5} \pm 3.9\mathrm{e}{-4}$    &    $\mathbf{3.0\mathrm{e}{-8} \pm 7.2\mathrm{e}{-8}}$    &
$2.1\mathrm{e}{+1} \pm 1.9\mathrm{e}{+3}$    &    $3.59\text{s} \pm 2.18\text{s}$ \\
HNN & Structured Hamiltonian & 
$\mathbf{4.6\mathrm{e}{-7} \pm 1.9\mathrm{e}{-6}}$    &    $\mathbf{4.6\mathrm{e}{-9} \pm 2.8\mathrm{e}{-8}}$    &     
$\mathbf{8.1\mathrm{e}{-8} \pm 5.3\mathrm{e}{-7}}$    &    $\mathbf{3.5\mathrm{e}{-8} \pm 8.3\mathrm{e}{-8}}$    &
$\mathbf{1.1\mathrm{e}{-4} \pm 4.4\mathrm{e}{-4}}$    &    $\mathbf{5.09\text{s} \pm 1.91\text{s}}$ \\
HNN & Black-Box Hamiltonian & 
$3.3\mathrm{e}{-5} \pm 5.5\mathrm{e}{-4}$    &     $9.9\mathrm{e}{-6} \pm 6.0\mathrm{e}{-5}$    &     
$3.3\mathrm{e}{-5} \pm 3.1\mathrm{e}{-4}$    &     $\mathbf{5.9\mathrm{e}{-8} \pm 1.4\mathrm{e}{-7}}$  &
$2.0\mathrm{e}{-2} \pm 3.1\mathrm{e}{-1}$    &    $3.80\text{s} \pm 1.72\text{s}$ \\
FF-NN & Feed Forward Network & 
$5.8\mathrm{e}{-5} \pm 1.0\mathrm{e}{-3}$    &     $2.3\mathrm{e}{-7} \pm 1.5\mathrm{e}{-6}$    &     
$9.9\mathrm{e}{-6} \pm 1.4\mathrm{e}{-4}$    &     $2.9\mathrm{e}{-7} \pm 7.4\mathrm{e}{-7}$    &
$6.1\mathrm{e}{-3} \pm 1.1\mathrm{e}{-1}$    &    $2.52\text{s} \pm 0.56\text{s}$ \\
\cmidrule(lr){1-2} \cmidrule(lr){3-3} \cmidrule(lr){4-4} \cmidrule(lr){5-5} \cmidrule(lr){6-6} \cmidrule(lr){7-7} \cmidrule(lr){8-8} 
\multicolumn{2}{c|}{\textbf{Character Data - $\#$ Samples} $\mathbf{= 2500}$} \\
\cmidrule(lr){1-2} \cmidrule(lr){3-3} \cmidrule(lr){4-4} \cmidrule(lr){5-5} \cmidrule(lr){6-6} \cmidrule(lr){7-7} \cmidrule(lr){8-8} 
DeLaN & Structured Lagrangian & 
$\mathbf{7.7\mathrm{e}{-8} \pm 2.1\mathrm{e}{-7}}$    &     $2.0\mathrm{e}{-7} \pm 1.7\mathrm{e}{-6}$    &     
$\mathbf{1.1\mathrm{e}{-6} \pm 5.2\mathrm{e}{-6}}$    &     $2.0\mathrm{e}{-7} \pm 1.0\mathrm{e}{-6}$    &
$\mathbf{3.5\mathrm{e}{-5} \pm 1.1\mathrm{e}{-4}}$    &    $\mathbf{2.32\text{s} \pm 0.37\text{s}}$ \\
DeLaN & Black-Box Lagrangian & 
$\mathbf{9.9\mathrm{e}{-8} \pm 4.4\mathrm{e}{-7}}$    &     $\mathbf{9.4\mathrm{e}{-8} \pm 4.7\mathrm{e}{-7}}$    &     
$\mathbf{1.7\mathrm{e}{-6} \pm 1.2\mathrm{e}{-5}}$    &     $1.3\mathrm{e}{-7} \pm 1.1\mathrm{e}{-6}$    &
$\mathbf{4.3\mathrm{e}{-5} \pm 1.7\mathrm{e}{-4}}$    &    $\mathbf{2.76\text{s} \pm 0.68\text{s}}$ \\
HNN & Structured Hamiltonian & 
$\mathbf{1.8\mathrm{e}{-8} \pm 5.5\mathrm{e}{-8}}$    &    $\mathbf{1.5\mathrm{e}{-8} \pm 9.1\mathrm{e}{-8}}$    &     
$\mathbf{5.5\mathrm{e}{-7} \pm 1.9\mathrm{e}{-6}}$    &    $\mathbf{2.5\mathrm{e}{-8} \pm 7.0\mathrm{e}{-8}}$    &
$\mathbf{1.5\mathrm{e}{-5} \pm 3.6\mathrm{e}{-5}}$    &    $\mathbf{2.90\text{s} \pm 0.68\text{s}}$ \\
HNN & Black-Box Hamiltonian & 
$\mathbf{5.5\mathrm{e}{-8} \pm 1.8\mathrm{e}{-7}}$    &     $5.0\mathrm{e}{-5} \pm 2.7\mathrm{e}{-4}$    &     
$7.2\mathrm{e}{-4} \pm 4.3\mathrm{e}{-3}$    &     $\mathbf{8.4\mathrm{e}{-8} \pm 2.7\mathrm{e}{-7}}$    &
$6.1\mathrm{e}{-5} \pm 1.6\mathrm{e}{-4}$    &    $2.19\text{s} \pm 0.62\text{s}$ \\
FF-NN & Feed Forward Network & 
$2.2\mathrm{e}{-6} \pm 9.1\mathrm{e}{-6}$    &     $4.2\mathrm{e}{-6} \pm 2.5\mathrm{e}{-5}$    &     
$5.5\mathrm{e}{-5} \pm 3.2\mathrm{e}{-4}$    &     $9.2\mathrm{e}{-7} \pm 5.1\mathrm{e}{-6}$    &
$6.5\mathrm{e}{-4} \pm 3.0\mathrm{e}{-3}$    &    $1.81\text{s} \pm 0.49\text{s}$ \\
\bottomrule
\end{tabular*}
\label{table:results}
\end{table*}

\medskip \noindent
\textbf{Feed-Forward Network.} The deep network baseline uses two separate networks, where one describes the forward dynamics and the other the inverse dynamics. This model does not necessarily generate coherent predictions as the parameters of the forward and inverse model are decoupled. Therefore, it is not guaranteed that $f(f^{-1}\left(\vx)\right) = \vx$ holds. The forward model is a continuous-time model and predicts the joint acceleration $\ddot{\vq}$. Therefore, the deep network baseline is independent of the sampling frequency and uses an explicit integrator as the physics-inspired network. The network parameters are learned by minimizing the normalized squared error of the forward and inverse model. This optimization problem is solved by gradient descent using ADAM. This baseline cannot be applied to the energy experiments, as the system energy cannot be learned by a standard deep network.

\subsection{Model Prediction Experiments}
For the simulated experiments, we want to evaluate whether the physics-inspired networks can learn the underlying system dynamics and recover the structure with ideal observations. Therefore, we want to observe the data fit and as well as the long-term forward predictions. \textcolor{revision}{Furthermore, we want to differentiate between two separate datasets, (1) a large data set with $100$k samples spanning the state domain uniformly and (2) a small dataset with only $2.5$k samples which consist of trajectories of drawing characters. For the character test dataset, the test set contains different characters than the training set.} This dataset only spans a small sub-domain of the state space. The character dataset was initially introduced by \cite{williams2008modelling} and is available in the UCI Machine Learning Repository \citep{Dua:2017}. For training, the datasets are split into a test and training set. The reported results are reported on the test set and averaged over 5 seeds.

\subsubsection{\textbf{Inverse Model.}} The results of the inverse model are summarized in Table \ref{table:results} and visualized in Figure~\ref{fig:inverse_model}. All models learn a good inverse model that fits the test set. When comparing the performance across the large and small datasets one cannot observe a difference in performance. All models perform comparably for the small and large datasets. On average, the physics-inspired networks obtain a lower MSE than the black-box deep network. When comparing the structured Lagrangian / Hamiltonian to the black-box counterparts no clear difference is observable for the inverse model. 

\smallskip\noindent
When comparing the torque decomposition of the inertial, centrifugal, Coriolis, and gravitational forces, all models learn a good decomposition. For the unstructured models, this decomposition can be evaluated by assuming the underlying structure and evaluating the inverse model. For example, the gravitational component by evaluating $\vtau_{g} = f^{-1}(\vq, \mathbf{0}, \mathbf{0})$. All models learn the underlying structure that fits the true decomposition. Even the black-box feed-forward network obtains a good decomposition despite having no structure. When comparing the MSE error in Table \ref{table:results}, the MSE for the physics-inspired networks is better than the black-box feed-forward network. This difference is especially pronounced for the inertial and Coriolis torque. The difference in the gravitational torque is not so large. When comparing the decomposition of the black-box Lagrangian / Hamiltonian to the structured counterparts, the structured approach outperforms the black-box approach on the inertial and Coriolis torque. 

\subsubsection{\textbf{Forward Model.}} The results of the forward model are summarized in Table \ref{table:results} and visualized in Figure~\ref{fig:forward_model}. Also for the forward model, the physics-inspired networks obtain a better performance on the state error than the feed-forward network. All models perform better on the small character dataset than on the large dataset as the state domain is much smaller than the uniform domain of the large dataset. For the large dataset, the black-box Lagrangian / Hamiltonian approaches are much worse compared to the structured counterparts. This is especially visible for the black-box Lagrangian. The average error and variance is so large because the mass matrix becomes nearly singular for some samples. The nearly singular mass matrix amplifies small differences yielding a very large error. 

\begin{figure*}[t]
    \centering
    \includegraphics[width=\textwidth]{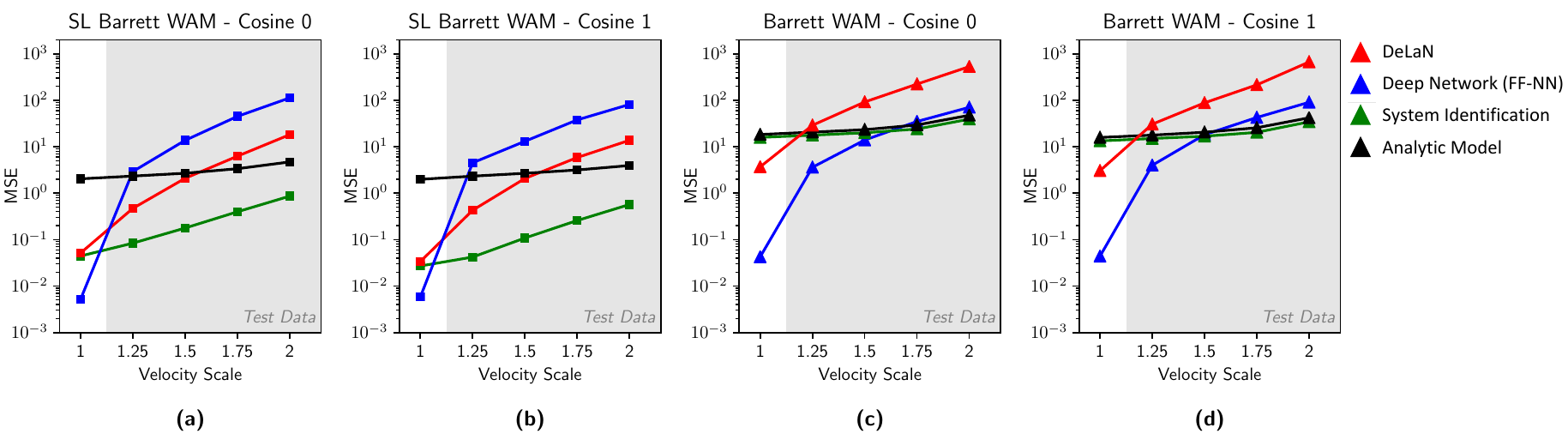}
    \caption{\textcolor{revision}{(a, b) The mean squared tracking error of the inverse dynamics control following cosine trajectories for the simulated Barrett WAM. (c, d) the mean squared tracking error on the physical Barrett WAM. The system identification approach, feed-forward neural network, and DeLaN are trained offline using only the trajectories at a velocity scale of 1×. Afterward, the models are tested on the same trajectories with increased velocities to evaluate the extrapolation to new velocities.}}
    \label{fig:inverse_dynamics_control}
\end{figure*}  

\medskip\noindent
To compare the long-term predictions of the models, we compare the \acf{vpt}~\citep{botev2021priors}, which is defined as the duration until the predicted rollout has a larger error than a pre-defined threshold. We define the threshold of the MSE to be $1\mathrm{e}{-2}$, which corresponds to an angular error of $\approx 5$ degrees. The long-term prediction of the physics-inspired networks is better than the prediction time of the feed-forward network on both datasets (Table \ref{table:results}). Furthermore, the structured variants of \ac{delan} and \ac{hnn} perform better than the black-box approaches. The problem of the nearly singular mass matrix can be observed in Figure~\ref{fig:forward_model} for the black-box Lagrangian. For one test trajectory and some seeds, the near singular mass matrix lets the trajectory diverge. For the structured \ac{hnn} and \ac{delan} this divergence is not observed. Furthermore, the momentum prediction of the black-box \ac{delan} variant shows the worse accuracy of the network Hessian corresponding to the mass matrix. The predicted momentum of this approach has a much higher variance.

\subsubsection{\textbf{Conclusion.}}
The simulated experiments show that the physics-inspired networks learn the underlying structure of the dynamical system. These models can accurately predict the force decomposition, momentum, and system energy. Furthermore, the physics-inspired models can learn better forward and inverse models than a standard feed-forward deep network. The structured \ac{delan} and \ac{hnn} perform better than the black-box counterparts. The forward and inverse model of the structured \ac{delan} and \ac{hnn} do not show any empirical differences when the corresponding phase space coordinates are observed.

\subsection{Model-Based Control Experiments}
With the experiments on the physical system, we want to evaluate the control performance of the learned models with noisy real-world data. Evaluating the control performance rather than the MSE on static datasets is the more relevant performance measure as the application of the models is control. Furthermore, it has been shown that the MSE is not a good substitute to predict the control performance of a learned model and commonly overestimates the performance \citep{hobbs1989optimization, lambert2020objective, lutter2021learning}. To evaluate the model performance for control, we apply the learned models to inverse dynamics control and energy control. We only apply \ac{delan} with the structured Lagrangian to the physical systems as the potentially singular mass matrix risks damaging the physical system. \ac{hnn} do not apply to the system as the momentum cannot be retrieved from the position observations while the velocity can be obtained using finite differences. 

\begin{figure*}[t]
    \centering
    \includegraphics[width=\textwidth]{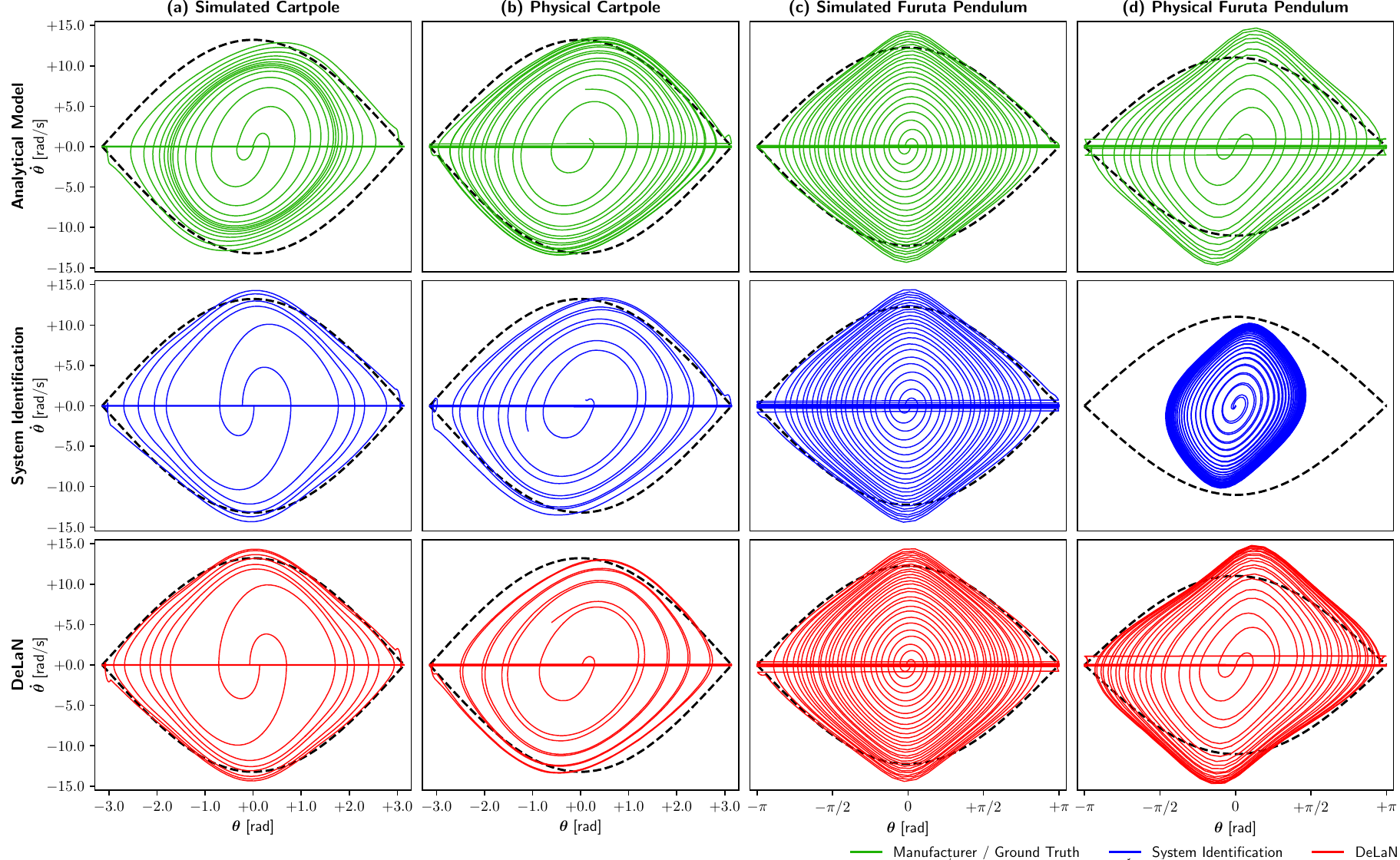} 
    \caption{The position $\theta$ and velocity $\dot{\theta}$ orbits recorded using energy control to swing up the cartpole and Furuta pendulum. The rows show the different models, i.e., the analytic model, the system identification model, and the \ac{delan} model while the columns show the different simulated and physical systems. The dashed orbit highlights the desired energy $E^{*}$. While the learned and the analytic model can swing up the simulated system and physical Cartpole only the analytic model and \ac{delan} can swing up the physical Furuta pendulum, while the energy controller using the System Identification model cannot.}
    \label{fig:energy_control}
\end{figure*}

\subsubsection{\textbf{Inverse Dynamics Control.}}
Evaluating learned models by comparing the tracking error of a model-based control law has been a well-established benchmark for evaluating the control performance of models \citep{nguyen2008computed, nguyen2009model}. In this experiment, we use inverse dynamics control as a model-based control law. This feedback controller augments the PD-control law with an additional feed-forward torque to compensate for the non-linear dynamics of the system. Therefore, the inverse dynamics control obtains a better tracking error than the standard PD control. The resulting control law is described by
\begin{gather*}
    \vtau = \mK_{p} \left(\vq_{\text{des}} - \vq \right) + \mK_{D} \left( \dot{\vq}_{\text{des}} - \dot{\vq} \right) + f^{-1}(\vq_{\text{des}}, \dot{\vq}_{\text{des}}, \ddot{\vq}_{\text{des}}),
\end{gather*}
with the position and derivative gains $\mK_p$ and $\mK_{d}$. In addition, we test the generalization of the learned models by increasing the velocity of the test trajectories. One would expect that \ac{delan} would generalize better to scaled velocities as the predicted mass matrix and potential energy only depend on the joint position and are independent of the velocity. \textcolor{revision}{The training and testing sequences consist of cosine trajectories with different frequencies for each joint and include a little chirp to avoid learning the Fourier basis. The test and train data differ in the frequency of each joint}. These trajectories are the standard approach to excite the system and cover a large state domain. The analytic model of the Barrett WAM is obtained from the \cite{JohnHopkinsUrdf}.

\medskip\noindent
The results for the simulated and physical Barrett WAM are summarized in Figure~\ref{fig:inverse_dynamics_control}. In the simulation, \ac{delan} and the system identification perform equally well on the training velocity. When comparing generalization, the system identification approach generalizes better than \ac{delan} to higher velocities. This behavior is expected as system identification obtains the global system parameter while \ac{delan} only learns a local approximation of the mass matrix and potential energy. In comparison to the feed-forward deep network, \ac{delan} performs worse on the training velocity but generalizes better to higher velocities. Therefore, the deep network overfits to the training velocity. The analytic model and the system identification have a large performance gap in simulation as we use the same analytic model for simulation and the physical system but the analytic model is optimized for the physical system.

\medskip\noindent
On the physical system, the feed-forward network performs the best on the training domain but deteriorates when the velocity is increased. \ac{delan} performs worse than the deep network but better than the analytic model and the system identification. The analytic model and the system identification model perform nearly identical. The system identification approach is only marginally better. Both approaches generalize better compared to \ac{delan} and the deep network. This better generalization is expected as the system parameters are global while the other approaches use local approximations. When increasing the velocity, \ac{delan} and deep network dynamics model degrade in performance. In contrast to the simulation results, where DeLaN extrapolates better than the deep network, the black-box deep network obtains the better generalization on the physical system. The worse performance and generalization of \ac{delan} on this physical system can be explained by the assumption of rigid body dynamics. This assumption is not fulfilled due to the cable drives and the motor-side sensing. Therefore, \ac{delan} cannot model every phenomenon with high fidelity. However, \ac{delan} learns a good approximation that is better than the system identification approach with the same rigid body assumption.

\subsubsection{\textbf{Energy Control Control.}}
A different approach to test the control performance of the learned models is to apply the learned models to controlling under-actuated systems using an energy controller. More specifically we apply an energy controller to swing up the Furuta pendulum and the cartpole. This energy controller regulates the system energy rather than the position and velocities. The control law is described by
\begin{gather*}
    \vu = k_{E}  \big[E(\vq, \dot{\vq}) - E(\vq_{\text{des}}, \dot{\vq}_{\text{des}}) \big] \: \sign\big( \dot{\vq}_{1} \cos(\vq_{1}) \big) - k_p \vq_1,
\end{gather*}
with the energy gain $k_{E}$ and position gain $k_p$. We use an additional position controller to prevent the system from hitting the joint limits. The control gains are tuned w.r.t. to the analytic model and fixed for all models. This control task is challenging as the control law relies on the system energy, which cannot be learned supervised. Therefore, the feed-forward network baseline cannot be applied to this task. In contrast to the feed-forward network, the physics-inspired deep network models are the first network models that can be applied to this task as these models can infer the system energy. \textcolor{revision}{For the training dataset, we use the energy controller to swing-up the pendulum, stabilize the pendulum at the top and let the pendulum fall down after $2$s. Once the pendulum settles the process repeated until about $200$s of data is collected.}

\medskip\noindent
The results for the simulated and physical experiments are summarized in Figure~\ref{fig:energy_control}. Videos of the physical experiments are available at \href{https://www.youtube.com/watch?v=m3JRYq7Gmgo}{[Link]}. Within the simulation, the analytic model, the system identification model, and \ac{delan} achieve the successful swing-up of the cartpole and Furuta pendulum. On the physical cartpole, all approaches achieve the swing-up despite the large stiction of the linear actuator. For the physical Furuta pendulum, only the analytic model and \ac{delan} achieve the swing-up. The system identification model does not. The system identification model fails as the linear regression is very sensitive to the observation noise and the small condition number of the features $\mA$ due to the small dimensions. Therefore, minor changes in the observation can lead to vastly different system parameters. In this specific case, the system identification approach underestimates the masses and hence, exerts too little action to swing up the pendulum and is stuck on the limit cycle.

\subsubsection{\textbf{Conclusion.}}
The non-linear control experiments on the physical systems show that \ac{delan} with a structured Lagrangian can learn a good model despite the noisy observations. The resulting model can be used for closed-loop feedback control in real-time for fully-actuated and under-actuated systems. For both systems categories \ac{delan} achieves a good control performance. It is noteworthy that \ac{delan} is the first model learning approach utilizing no prior knowledge of the equations of motion that can be applied to energy control. The previous black-box model learning approach could not be applied as the system energy can only be learned unsupervised.

\section{Conclusion}\label{sec:discussion} 
%
%

\medskip\noindent \textcolor{revision}{
Coming back to the initial questions of the experiments. The experimental results have showed that}

\medskip\noindent \textcolor{revision}{
\textbf{Q1:} physics-inspired networks can learn the underlying representation of the dynamical system unsupervised. The predicted inertial, centrifugal, Coriolis and gravitational forces match the ground-truth (Fig \ref{fig:inverse_model}). On the physical systems the learned system energy can be used for energy control. The swing-up of under-actuated Furuta pendulum and cartpole where successfully achieved.}

\medskip\noindent \textcolor{revision}{
\textbf{Q2:} physics-inspired networks may learn better models than continuous-time black-box models with feed-forward networks. Especially, in simulation the physics-inspired networks achieve lower mean squared error (Fig. \ref{fig:forward_model}), longer valid prediction times (Table \ref{table:results}) and better generalization (Fig. \ref{fig:inverse_dynamics_control}). On the physical system, when the assumptions of the physics-inspired networks are violated, the results are not as clear. While \ac{delan} performs comparable to system identification techniques, the deep network model can represent physical phenomena that are not captured in the physics prior and achieves a lower tracking error.}

\medskip\noindent \textcolor{revision}{
\textbf{Q3:} physics-inspired networks can be applied to physical systems. Even though the physical systems violate the physics prior due to non-ideal torque sources and unmodeled phenomena such as cable drives, motor dynamics and backslash, \ac{delan} was able to learn the system energy. For example, the energy controller utilizing \ac{delan} was able to solve the swing-up the Furuta pendulum and the cartpole, which has large backslash in the linear actuator.} 

\medskip\noindent \textcolor{revision}{
Similar empirical results were also presented by \citep{greydanus2019hamiltonian, cranmer2020lagrangian, gupta2019general, gupta2020structured, saemundsson2020variational, zhong2019symplectic, zhong2020dissipative}. When comparing the different physics prior, Hamiltonian and Lagrangian priors yield comparable models when the corresponding phase space coordinates are observed, i.e., velocities for \ac{delan} and impulse for \ac{hnn}. When comparing structured \ac{delan} and \ac{hnn} to the black-box \ac{delan} and \ac{hnn}, we find that the structured approaches achieve better dynamics models (Table \ref{table:results}). The black-box variants struggle to obtain non-singular network Hessians for all possible system states. For the structured approaches singular Hessians can be prevented by parametrized the kinetic energy such that the eigenvalues of the Hessian are lower-bounded and do not become singular.}

\medskip\noindent \textcolor{revision}{
Despite the advantages of the physics-inspired models to standard deep networks models, the physics-inspired approaches have drawbacks that prevent the general applicability compared to feed-forward networks. In the following, we discuss these limitations.}

\subsection{Open Challenges}
Physics-inspired deep networks have two main shortcomings, which have not been solved yet. First of all, the current approaches are only able to simulate articulated rigid-bodies \emph{without} contact and second the current approaches rely on knowing and observing the generalized coordinates. Therefore, most of the existing work only showcased these networks for simple n-link pendulums \citep{zhong2021differentiable} and n-body problems \citep{sanchezgonzalez2019hamiltonian}. For most real-world robotic tasks these assumptions are not full-filled. One frequently does not know or observe the system state and most interesting robotic tasks include contacts. In contrast to physics-inspired networks, black-box dynamics models work with any observations and contacts. These models have been extensively used for model predictive control and are sufficient for complex control tasks~\citep{hafner2019dream, hafner2019learning, lutter2021learning}. Therefore, these challenges need to be addressed to enable the widespread use of physics-inspired methods for robot control. 
In the following, we highlight the challenges of both limitation and the initial step towards applying these models to contact-rich tasks with arbitrary observations.

\subsubsection{\textbf{Contacts.}}
Analytically, contact forces can be incorporated by adding generalized contact forces to the Euler-Lagrange equation. In this case the differential equation is described by
\begin{align*}
    \frac{d}{dt} \frac{\partial \mathcal{L}(\vq, \dot{\vq})}{\partial \dot{\vq}} - \frac{\partial \mathcal{L}(\vq, \dot{\vq})}{\partial \vq} &= \bm{\tau} + \underbrace{\sum_{i \in \, \Omega} \mJ^{c}_i(\vq) \vf^{c}_i(\vq, \dot{\vq})}_{\text{Generalized Contact Force}},
\end{align*}
with the Cartesian contact forces~$\vf^c$, the contact Jacobian~$\mJ^{c}$ connecting the Cartesian forces at the contact point to the generalized coordinates and the set of all active contacts~$\Omega$. To compute the generalized contact force, analytic simulators first use the known kinematics and meshes to find all contact points and there respective Jacobians. Afterwards the contact force is computed by solving the linear complementarity problem (LCP). A similar approach can also be used for Hamiltonian mechanics that uses contact impulses rather than forces.  
\medskip \noindent
Within the physics-inspired deep network literature only \cite{hochlehnert2021learning} and \cite{zhong2021differentiable} have included contacts. However, both existing works only consider special cases with strong assumptions. For example, \cite{hochlehnert2021learning} only considers elastic collisions of simple geometric shapes, i.e., circles. In this case, the contact forces can be computed and the contact Jacobian is the identity matrix. Therefore, one only needs to learn an indicator function $\vmathbb{1}(\vq)$ being $1$ if the contact is active and $0$ otherwise. Furthermore, the indicator function is learned supervised. Hence, the training data has to include whether the contact was active or not for each sample. The experiments only apply the proposed algorithm to a ball bouncing on a plane and the Newton cradle. 

\medskip \noindent A different approach was proposed by \cite{zhong2021differentiable}. This work augments the physics-inspired network with a differentiable physics simulator to handle the contacts. In this case, a collision detection algorithm determines all active contacts and the contact Jacobians. The contact forces are computed by solving the LCP. In this case, only the coefficients of the contact model, e.g., friction and restitution, are learned from data. Therefore, this approach is similar to the white-box friction models described in Section \ref{sec:friction}. This approach also implicitly assumes that the meshes and kinematics are \emph{known}. Without the kinematics and meshes the collision detection algorithms cannot compute the active contacts and Jacobians. If these quantities of the system are known, the analytic equations of motions can be computed and many physical parameters can be approximated from the meshes. Therefore, these assumptions are identical to the required knowledge for system identification using differentiable physics simulators~\citep{werling2021fast, degrave2019differentiable, heiden2021disect}. The advantage of physics-inspired networks compared to system identification with differentiable simulators are unknown. The experiments only applied the proposed approach to bouncing disks and a multi-link pendulum with a ground plane. 

\medskip \noindent 
To summarize, no general way to add contacts to physics-inspired networks has been proposed and shown to work for multi-contact physics with complex geometries. The naive approach to add a single network to model the generalized contact forces is challenging as this reduces the physics-inspired model learning approaches to a black-box model learning technique without proper regularization. Therefore, an important open challenge for physics-inspired networks for robotics is to introduce a generic approach to include multiple contacts.

\subsubsection{\textbf{Generalized Coordinates.}} \label{sec:generalized_coordinates}
The second limiting assumption is the observation of the generalized coordinates~$\vq, \dot{\vq}$ or the generalized momentum~$\vp$. For most robotic systems that do not only involve a rigid body manipulator, these coordinates are commonly not observed or known. One usually only obtains observations derived from the generalized coordinates if the system is fully observed. In many cases, the system is only partially observed and one cannot infer the system state from a single observation. For black-box models this is not a problem as these models do not require specific observations and have been shown to learn good dynamics models for complex systems using only images, e.g., \cite{hafner2019dream, hafner2019learning, hafner2020mastering} and many others.

\medskip \noindent
To overcome this limitation, existing work combined physics-inspired networks with variational autoencoders (VAE) to learn a latent space the resembles the generalized coordinates. In this case, the Lagrangian and Hamiltonian inspired networks are applied in the latent space. Using this approach, the dynamics of single-link pendulums and N-body problems have been learned from artificial images~\citep{greydanus2019hamiltonian, zhong2019symplectic, toth2019hamiltonian, saemundsson2020variational, allen2020lagnetvip}. However, these approaches have not been demonstrated on more complex systems and realistic rendering of systems. \cite{botev2021priors} also showed that this approach does not necessarily obtain better results than using a normal deep network continuous-time model within the latent space. Therefore, it remains an important open challenge to extend physics-inspired networks to arbitrary observations. The main challenge is to learn a latent space that resembles the generalized coordinates and the naive approach to use a VAE does not seem to be sufficient.

\vspace{-0.2em}
\subsection{Summary}
We introduced physics-inspired networks that combine Lagrangian and Hamiltonian mechanics with deep networks. This combination obtains physically plausible dynamics models that guarantee to conserve energy. The resulting models are also interpretable and can be used as forward, inverse, or energy models using the same parameters. Previously this was not possible with standard deep network dynamics models. Furthermore, we presented all the existing extensions of physics-inspired networks which include different representations of the Hamiltonian and Lagrangian, different loss functions as well as different actuation and friction models. We elaborated on the shortcomings of the current approaches as these techniques are limited to mechanical systems without contacts and require the observation of generalized positions, velocity, momentum, and forces. Therefore, this summary provides the big picture of physics-inspired networks for learning continuous-time dynamics models of rigid body systems.  

\medskip\noindent
Within the experiments, we showed that \acf{delan} and \acf{hnn} learn the underlying structure of the dynamical system for simulated and physical systems. When the corresponding phase-space coordinates of each model are observed, both models perform nearly identical. On average the structured Hamiltonian and Lagrangian perform better than their black-box counterparts. Especially for the Lagrangian combination, the black-box approach can lead to high prediction errors due to inverting the Hessian of a deep network. Furthermore, we show that these physics-inspired techniques can be applied to the physical system despite the observation noise. The resulting \ac{delan} models can be used for real-time control and achieve good performance for inverse dynamics control as well as energy control. Especially the latter is noteworthy, as \ac{delan} is the first model learning technique that utilizes deep networks and can learn the system energy. Previously this was only possible using system identification which requires knowledge of the kinematic structure to derive the equations of motion.  

\section*{Acknowledgments}
This project has received funding from ABB and NVIDIA. Furthermore, we want to thank the open-source projects NumPy~\citep{numpy}, PyTorch~\citep{pytorch} and JAX~\citep{jax2018github}. 

\vspace{-1em}
\bibliographystyle{sageH}
\bibliography{refs} 

\newpage
\onecolumn
\section{Appendix}
\subsection{Extended Experimental Results}
To make the differences between the different learned models shown in Figure \ref{fig:inverse_model} and Figure \ref{fig:forward_model} clearer, we provide a figure per model in the appendix. For more details about the figures please refer to the figures in the article and section \ref{sec:experiments}.

\begin{figure*}[h]
    \centering
    \includegraphics[width=\textwidth]{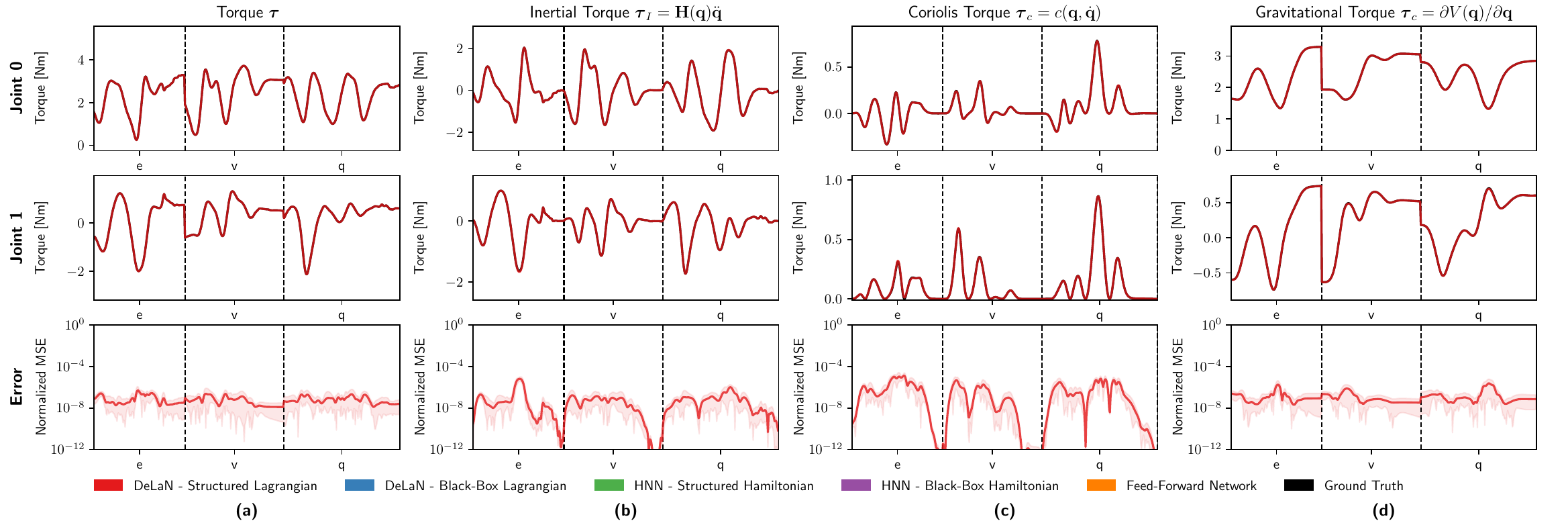}
    \caption{\textcolor{revision}{The learned inverse model of structured DeLaN using the character dataset averaged over 10 seeds. For more information see Fig. \ref{fig:inverse_model}.}}
\end{figure*} 

\begin{figure*}[h]
    \centering
    \includegraphics[width=\textwidth]{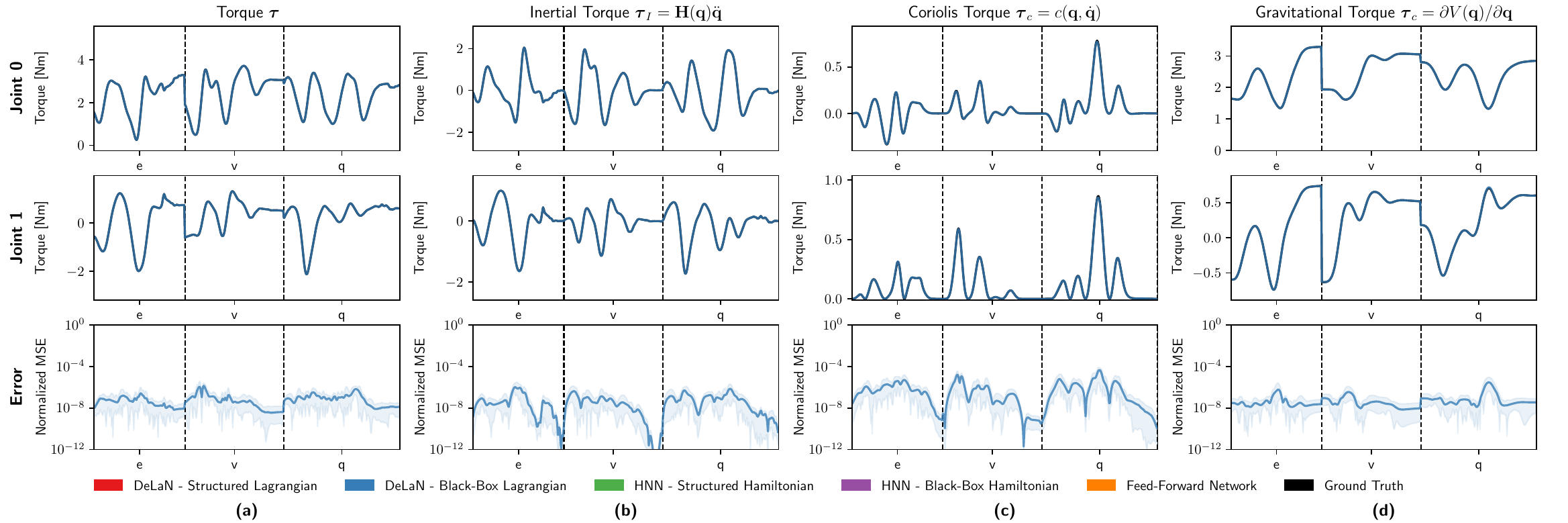}
    \caption{\textcolor{revision}{The learned inverse model of black-box DeLaN using the character dataset averaged over 10 seeds. For more information see Fig. \ref{fig:inverse_model}.}}
\end{figure*} 

\begin{figure*}[h]
    \centering
    \includegraphics[width=\textwidth]{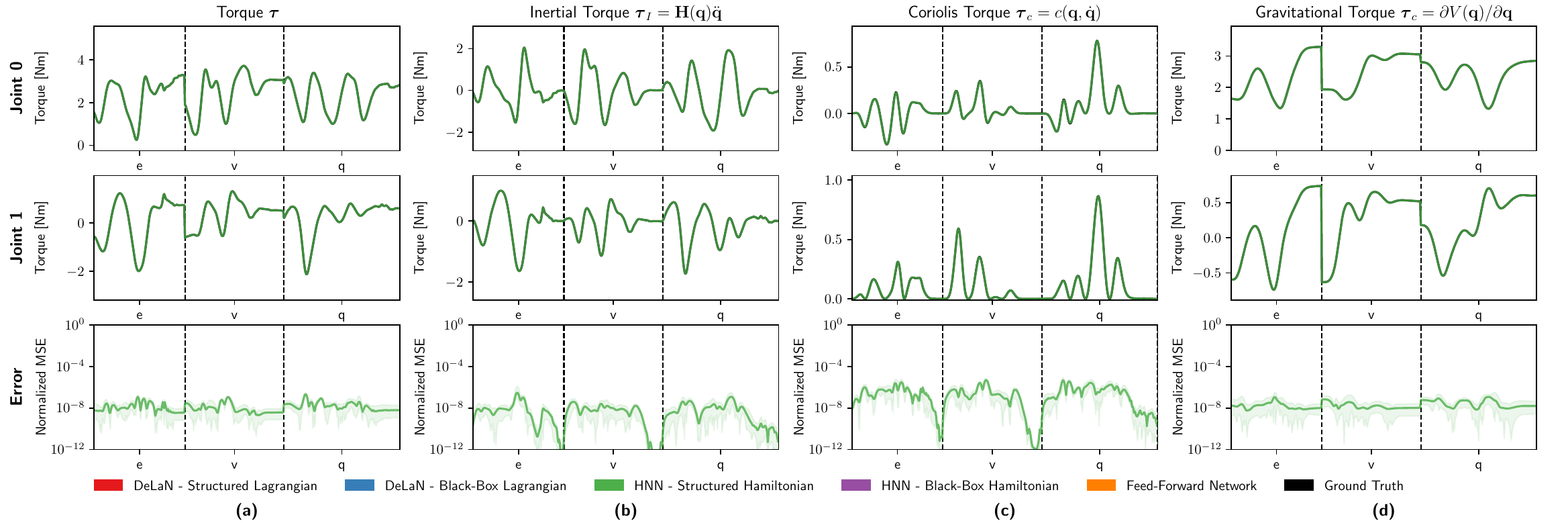}
    \caption{\textcolor{revision}{The learned inverse model of structured HNN using the character dataset averaged over 10 seeds. For more information see Fig. \ref{fig:inverse_model}.}}
\end{figure*} 

\begin{figure*}[h]
    \centering
    \includegraphics[width=\textwidth]{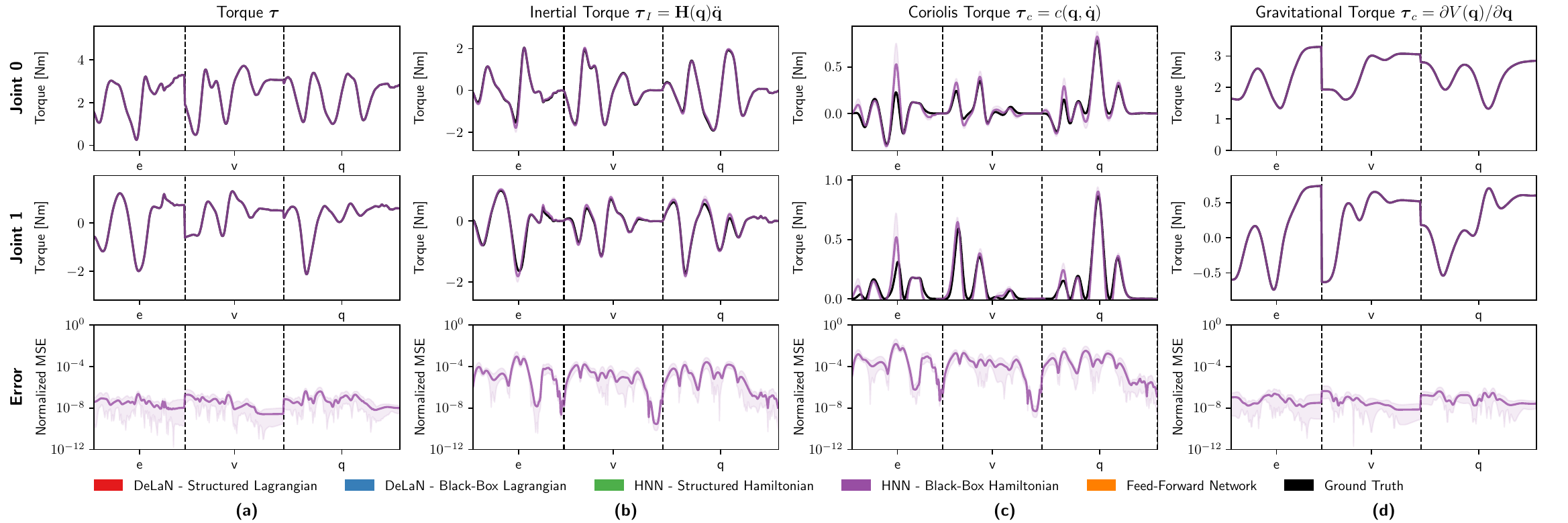}
    \caption{\textcolor{revision}{The learned inverse model of black-box HNN using the character dataset averaged over 10 seeds. For more information see Fig. \ref{fig:inverse_model}.}}
\end{figure*} 

\begin{figure*}[h]
    \centering
    \includegraphics[width=\textwidth]{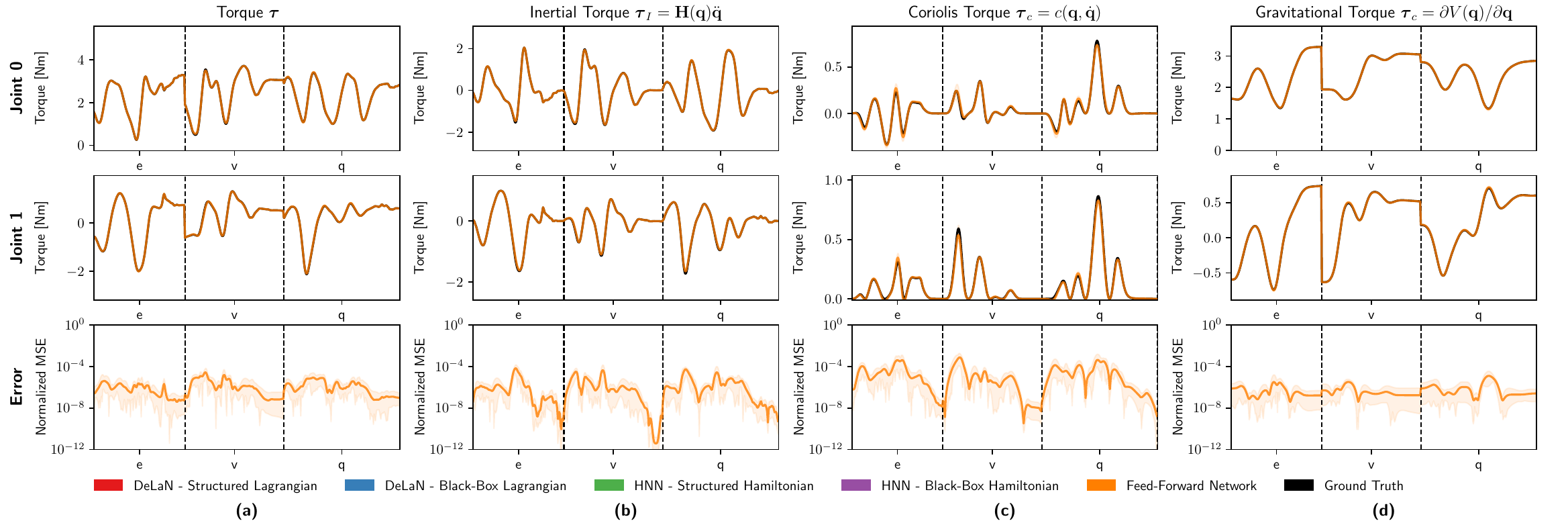}
    \caption{\textcolor{revision}{The learned inverse model of the feed-forward network using the character dataset averaged over 10 seeds. For more information see Fig. \ref{fig:inverse_model}.}}
\end{figure*} 

\begin{figure*}[h]
    \centering
    \includegraphics[width=\textwidth]{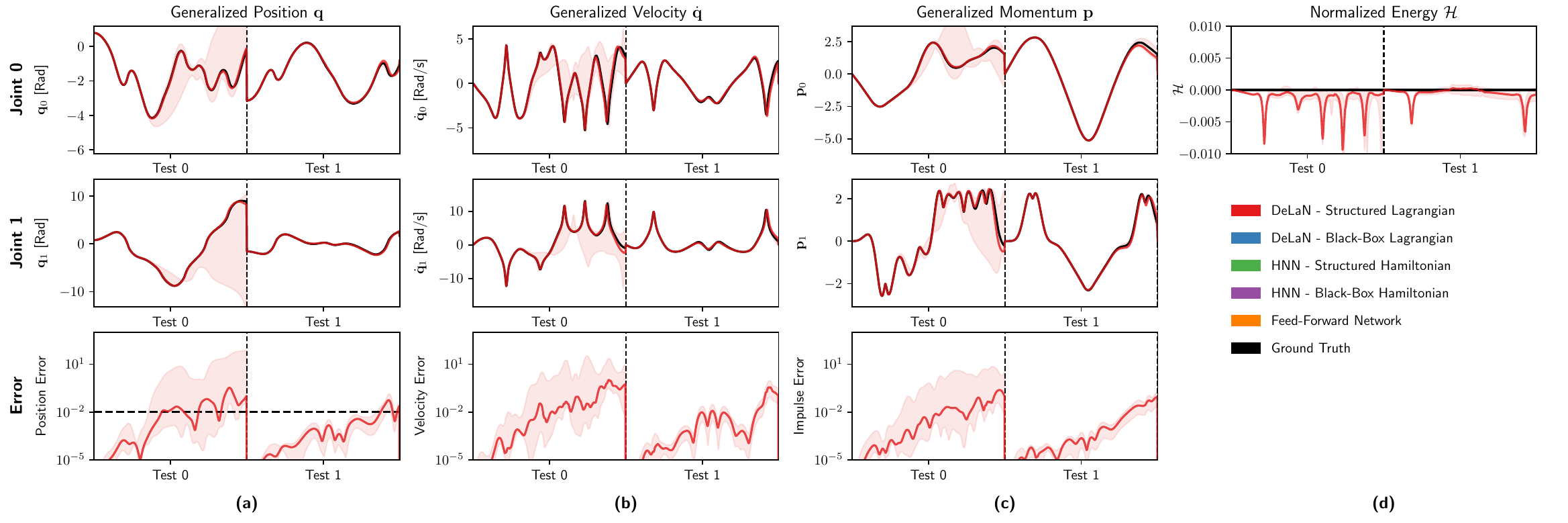}
    \caption{\textcolor{revision}{The model rollouts of (a) the position, (b) velocity and (c) momentum, and (d) energy of the structured DeLaN model trained on the uniform dataset averaged over 10 seeds. For more information see Fig. \ref{fig:forward_model}}}
\end{figure*} 

\begin{figure*}[h]
    \centering
    \includegraphics[width=\textwidth]{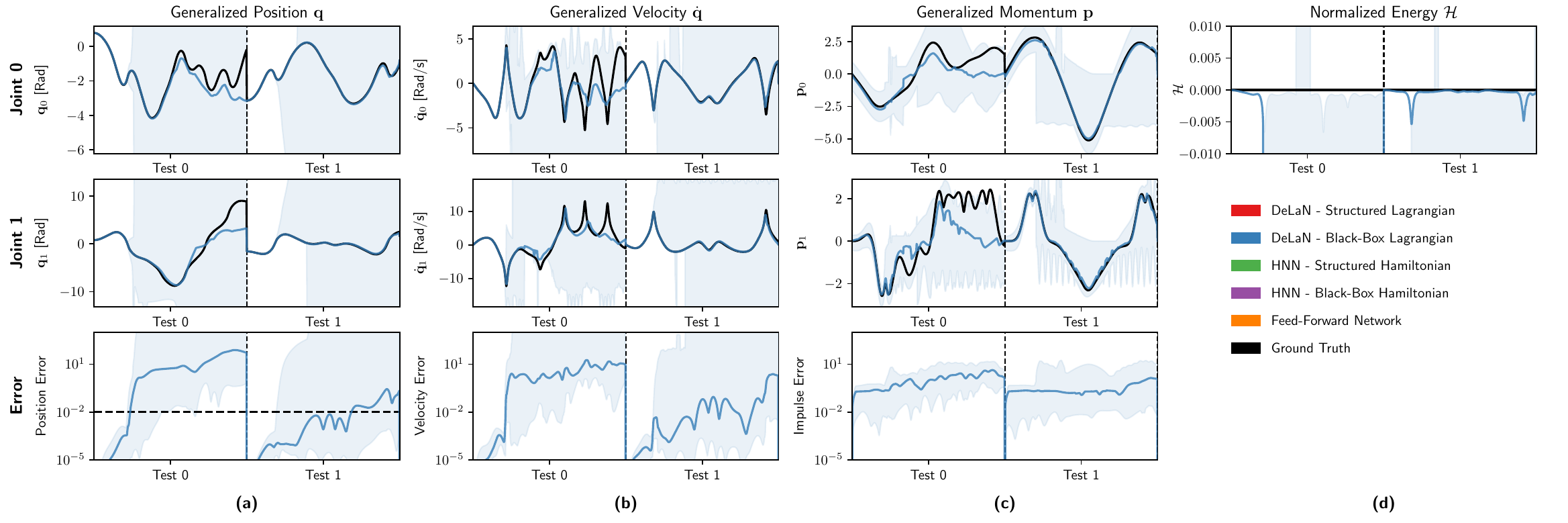}
    \caption{\textcolor{revision}{The model rollouts of (a) the position, (b) velocity and (c) momentum, and (d) energy of the black-box DeLaN model trained on the uniform dataset averaged over 10 seeds. For more information see Fig. \ref{fig:forward_model}}}
\end{figure*} 

\begin{figure*}[h]
    \centering
    \includegraphics[width=\textwidth]{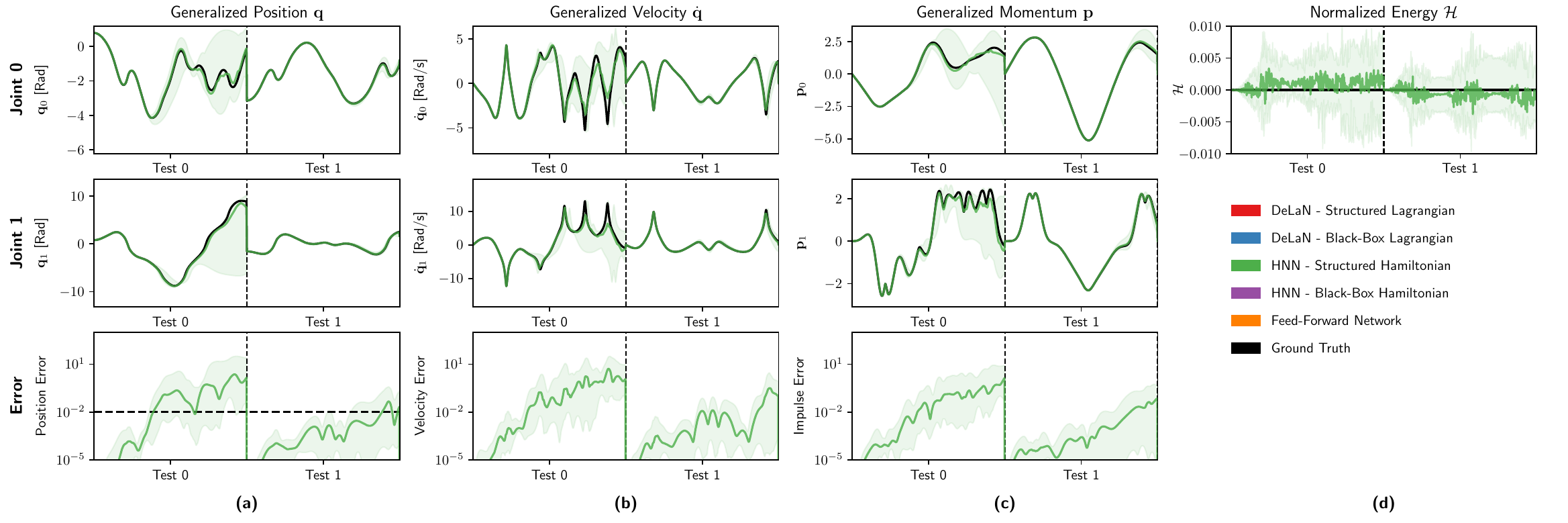}
    \caption{\textcolor{revision}{The model rollouts of (a) the position, (b) velocity and (c) momentum, and (d) energy of the structured HNN model trained on the uniform dataset averaged over 10 seeds. For more information see Fig. \ref{fig:forward_model}}}
\end{figure*} 

\begin{figure*}[h]
    \centering
    \includegraphics[width=\textwidth]{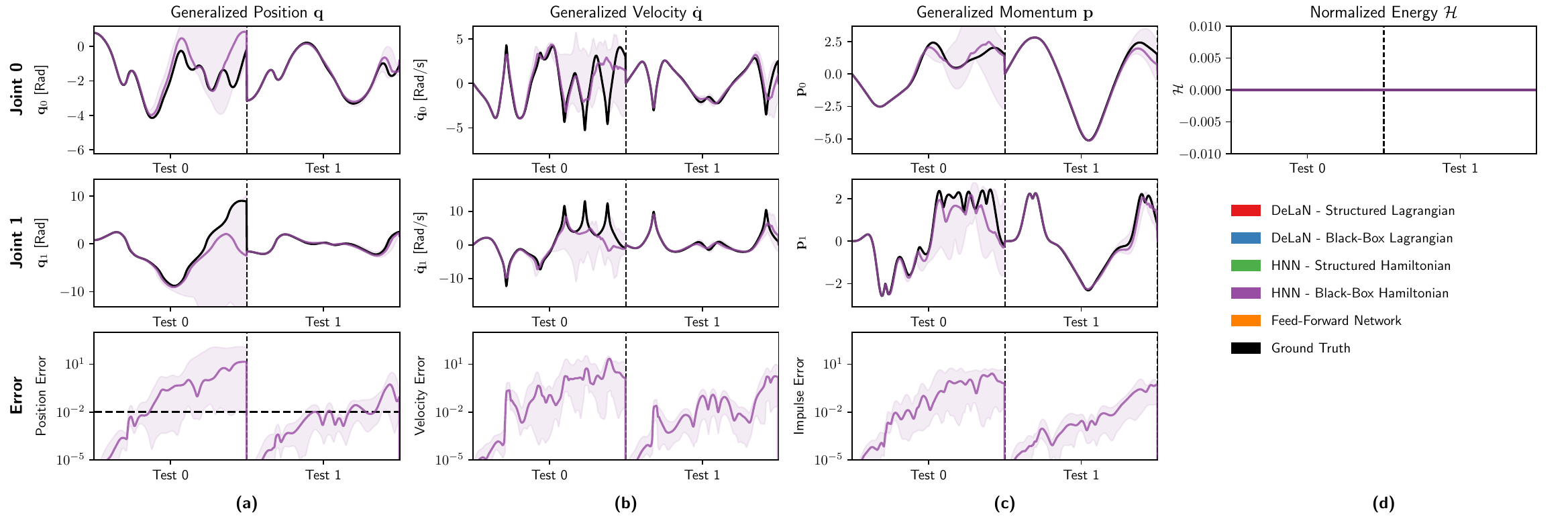}
    \caption{\textcolor{revision}{The model rollouts of (a) the position, (b) velocity and (c) momentum, and (d) energy of the black-box HNN model trained on the uniform dataset averaged over 10 seeds. For more information see Fig. \ref{fig:forward_model}}}
\end{figure*} 

\begin{figure*}[t]
    \centering
    \includegraphics[width=\textwidth]{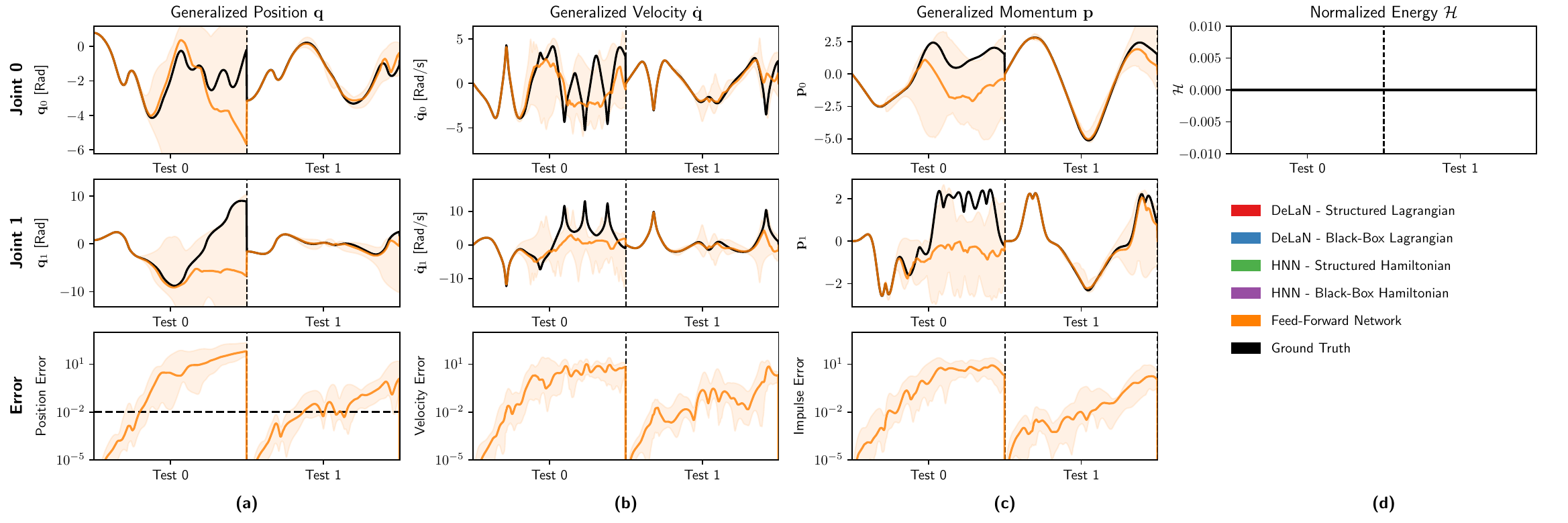}
    \caption{\textcolor{revision}{The model rollouts of (a) the position, (b) velocity and (c) momentum, and (d) energy of the feed-forward network model trained on the uniform dataset averaged over 10 seeds. For more information see Fig. \ref{fig:forward_model}}\vspace*{8in}}
\end{figure*} 
\newpage
\end{document}